\documentclass[10pt,twocolumn,letterpaper]{article}

\usepackage{iccv}
\usepackage{times}
\usepackage{epsfig}
\usepackage{graphicx}
\usepackage{amsmath}
\usepackage{amssymb}
\usepackage{threeparttable}
\usepackage{booktabs}
\usepackage{multirow}

\usepackage{makecell}
\usepackage{graphicx}
\usepackage[table]{xcolor}
\usepackage{arydshln}
\usepackage{marvosym}

\usepackage[accsupp]{axessibility}

\definecolor{blue}{RGB}{202,235,236}

\usepackage[pagebackref=true,breaklinks=true,letterpaper=true,colorlinks,bookmarks=false]{hyperref}

\iccvfinalcopy 

\ificcvfinal\fi

\begin{document}

\title{DetZero: Rethinking Offboard 3D Object Detection with Long-term Sequential Point Clouds}

\author{Tao Ma$^{1,2}$\thanks{Work performed during internship at Shanghai Artificial Intelligence Laboratory. \newline} \and
Xuemeng Yang$^{2}$ \and Hongbin Zhou$^{2}$ \and Xin Li$^{3,2*}$ \and Botian Shi$^{2}$ \and Junjie Liu$^{4*}$ \and Yuchen Yang$^{5,2*}$ \and Zhizheng Liu$^{6*}$ \and Liang He$^{3}$ \and Yu Qiao$^{2}$ \and Yikang Li$^{2}$\textsuperscript{\Letter} \and Hongsheng Li$^{1,2,7}$\textsuperscript{\Letter}
\\
[2mm]
$^1$Multimedia Laboratory, The Chinese University of Hong Kong \\  $^2$Shanghai Artificial Intelligence Laboratory \ \ $^3$East China Normal University \\
$^4$South China University of Technology \ \ $^5$Fudan University  \ \ $^6$ETH Zurich \ \ $^7$CPII \\
{\normalsize \Letter \ Corresponding author}\\
{\normalsize \url{https://github.com/PJLab-ADG/DetZero}}
}

\maketitle
\ificcvfinal\thispagestyle{empty}\fi

\begin{abstract}
Existing offboard 3D detectors always follow a modular pipeline design to take advantage of unlimited sequential point clouds. We have found that the full potential of offboard 3D detectors is not explored mainly due to two reasons: (1) the onboard multi-object tracker cannot generate sufficient complete object trajectories, and (2) the motion state of objects poses an inevitable challenge for the object-centric refining stage in leveraging the long-term temporal context representation. To tackle these problems, we propose a novel paradigm of offboard 3D object detection, named DetZero. Concretely, an offline tracker coupled with a multi-frame detector is proposed to focus on the completeness of generated object tracks. An attention-mechanism refining module is proposed to strengthen contextual information interaction across long-term sequential point clouds for object refining with decomposed regression methods.
Extensive experiments on Waymo Open Dataset show our DetZero outperforms all state-of-the-art onboard and offboard 3D detection methods. Notably, DetZero ranks 1st place on Waymo 3D object detection leaderboard\footnote{https://waymo.com/open/challenges/2020/3d-detection/.} with 85.15 mAPH (L2) detection performance.
Further experiments validate the application of taking the place of human labels with such high-quality results.
Our empirical study leads to rethinking conventions and interesting findings that can guide future research on offboard 3D object detection.
\end{abstract}

\section{Introduction}
Autonomous driving has rapidly advanced with promising progress in both industry and academia.
A crucial component of this development is offboard 3D object detection, which can utilize entire sequence data from sensors (video or sequential point cloud) with few constraints on model capacity and inference speed.
Therefore, some approaches~\cite{3dal,auto4d} are dedicated to developing high-quality ``auto labels'', aiming to reduce manual labor in point cloud annotation.

\begin{figure}[t]
    \centering
    \includegraphics[width=0.48\textwidth]{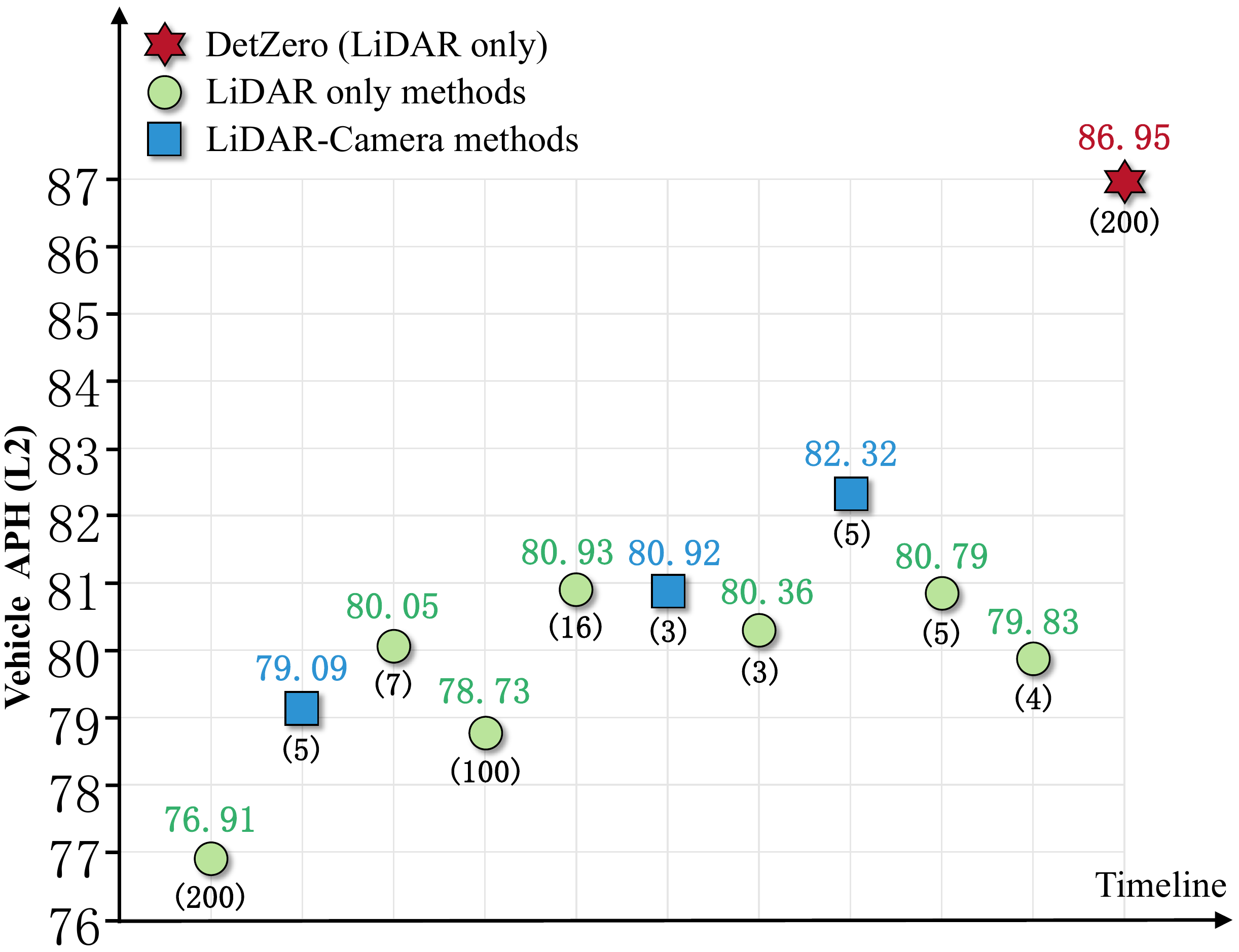}
    \caption{We compare several SOTA methods proposed along the timeline on the Waymo 3D detection leaderboard. Our DetZero obtains the best performance with a remarkable margin on \textit{Vehicle}. We mark the number of input point cloud frames below each method. Please refer to Table~\ref{table:leaderboard} for a detailed comparison.}
    \label{fig:leaderboard}
\end{figure}

Subsequently, many online detectors~\cite{3dman,mppnet,lidarmultinet} are introduced with the majority focusing on developing sophisticated modules to better utilize temporal context. As shown in Fig.~\ref{fig:leaderboard}, these newly proposed methods outperform both online~\cite{voxelnet,pointpillars,second,pointrcnn,centerpoint,pvrcnn,parta2,li2022homogeneous,kong2023robo3d,huang2022multi} and offboard 3D detectors~\cite{3dal,auto4d,int} by a large margin, leaving the impression that current architecture and pipeline of offboard 3D detectors are too weak to learn the complex representation over long-term sequential point clouds.
Therefore, in this paper, we revisit state-of-the-art (SOTA) offboard 3D detectors (see Sec.~\ref{sec:prelim} for the pipeline) and identify two main factors hindering the full potential: 
(1) the onboard multi-object tracker can't generate sufficient complete object trajectories,
and (2) the motion state of objects poses an inevitable challenge for the object-centric refining stage to leverage the long-term temporal context representation.
\begin{figure}[t]
    \centering
    \includegraphics[width=0.48\textwidth]{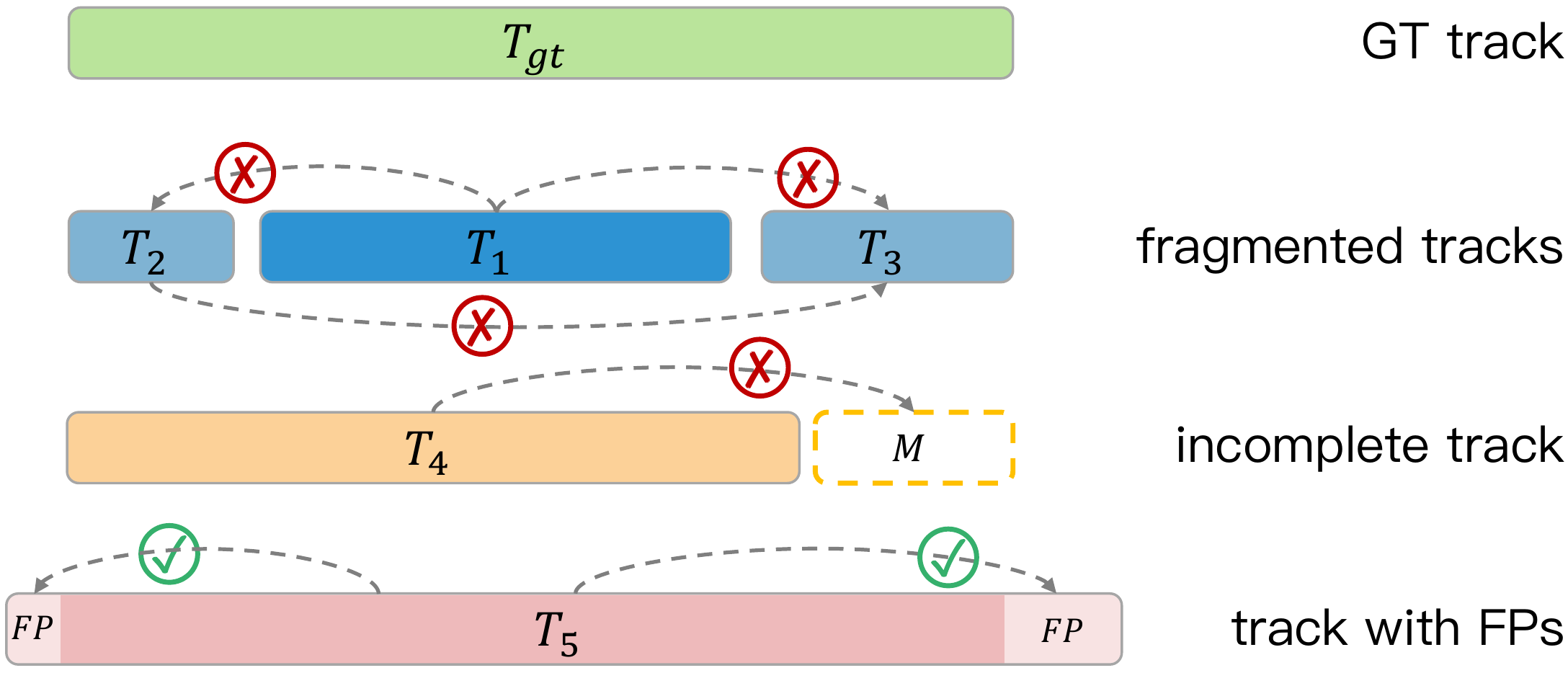}
    \caption{The quality of object tracks affects following refining much. Compared to $T^\text{gt}$, $T_1$, $T_2$, $T_3$ are fragmented tracks caused by ID switches, $T_4$ is an incomplete track with missing boxes $M$, $T_5$ contains many false positive (FP) boxes at both the beginning and the end. The dashed lines represent coordinate transform with the pose matrix. During object-centric refining, the full sequential context cannot be utilized across $T_1, T_2, T_3$; missing boxes $M$ cannot be compensated from existing recalled part of $T_4$;
    refined boxes with higher scores are still FP after being transformed to FP's location.}
    \label{fig:track-prob}
\end{figure}

\begin{figure}[t]
    \centering
    \includegraphics[width=0.48\textwidth]{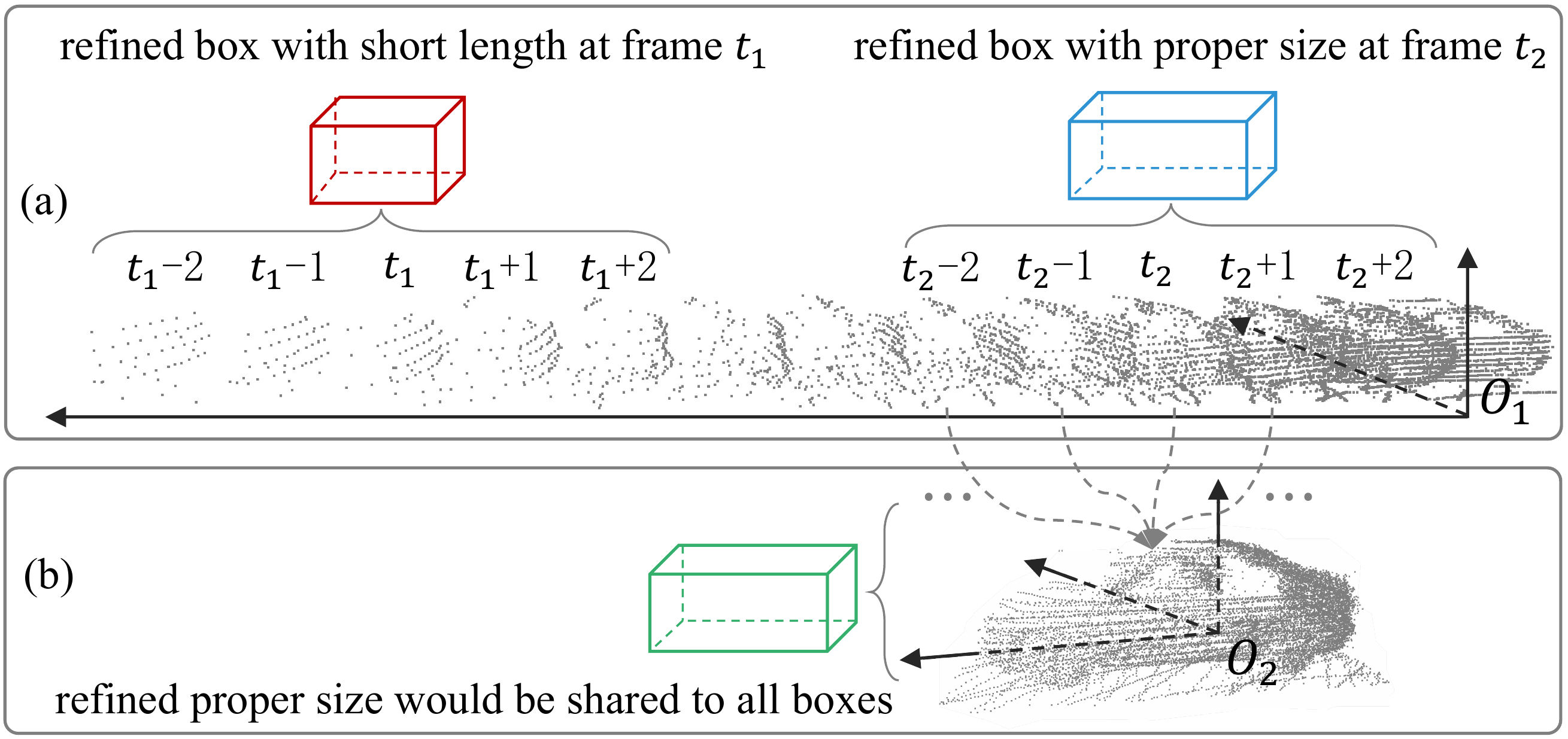}
    \caption{For a dynamic object track, the sliding-window based refining model (a) outputs inaccurate box size at frame $t_1$ when the object points are sparse, since it can't utilize dense point features at frame $t_2$. In (b), we merge object points together (from origin $O_1$ to $O_2$), then the points from each frame can contribute to precise size prediction.}
    \label{fig:refiner-prob}
\end{figure}

Specifically, prevailing online 3D detectors achieve promising performance but easily generate severe fragment trajectories, ID switches, and false positives when coupled with a tracking-by-detection multi-object tracking algorithm. As shown in Fig.~\ref{fig:track-prob}, this phenomenon may prevent the generation of complete temporal context features. Therefore, we adopt an upstream module comprising a multi-frame 3D detector and offline tracker that ensures the completeness and continuity of object tracking while maintaining high recall.
Moreover, the sliding-window-based auto labeling model~\cite{3dal, auto4d} hinders the exploitation of the commonality of object features, as shown in Fig.~\ref{fig:refiner-prob}. We notice that the size of the objects remained consistent over time. By capturing data from various viewpoints, we can enhance the point cloud of an object, allowing for more precise size estimation. Furthermore, the object trajectory is independent of its size, and should always follow the kinematic constraints in continuous time, which is manifested by the smoothness of the trajectory. These characteristics serve as the foundation for leveraging the long-term sequential point clouds in a decomposed paradigm: refine the geometry size, smooth the trajectory position, and update the confidence score.

By focusing on these main issues, we propose a new paradigm of offboard 3D object detection named DetZero. A tenet is underscored here: emphasizing high-recall detection and tracking during the upstream, meticulous high-accuracy refining with long-term temporal context during the downstream.
Comprehensive empirical studies and evaluations on the Waymo Open Dataset (WOD) demonstrate that DetZero significantly improves perception by fully utilizing long-term sequential point clouds. 
Notably, we rank $1$st place on WOD 3D object detection leaderboard with $85.15$ mAPH (L2). Extensive ablation studies and generalization experiments show that our method performs well with different quality upstream inputs and stricter metrics. Semi-supervised experiments further demonstrate that our method can provide high-quality auto labels for onboard 3D object detection models, which are already on par or even slightly higher than human labels.

The main contributions of our work are summarized as follows:
\begin{itemize}
    \item  We introduce DetZero, a new paradigm of offboard 3D object detection, to activate the potential of long-term sequential point clouds.
    \item Our proposed multi-frame object detection and offline tracking module generates accurate and complete object tracks, which is crucial for downstream refinement.
    \item An attention-mechanism based refining module is proposed to leverage the long-term temporal contextual information for objects' attribute predictions.
    \item We achieve state-of-the-art 3D object detection performance and tracking performance on the challenging WOD with remarkable margins.
\end{itemize}

\section{Related Work}
\noindent \textbf{3D Object Detection.}
Current 3D object detectors usually process the point cloud in different manners: grid-based and point-based. Different grid-split schemes are designed to transform the point cloud into 3D voxels~\cite{voxelnet,second,hvnet,centerpoint}, pillars~\cite{pillar,pointpillars} and bird-eye view maps~\cite{pixor} representation. Point-based methods~\cite{pointrcnn,parta2,std,3dssd,pointgnn,votenet} often employ PointNet~\cite{pointnet,pointpp} as a base feature extractor. The hybrid strategy~\cite{pvrcnn,mvf,std,sassd} is also utilized to leverage both advantages. Besides, transformer networks make a success to extract point clouds feature by attention mechanism~\cite{voxelformer,centerformer,swformer,ct3d,guan2021m3detr,conquer}, which have shown great potential.

\noindent \textbf{3D MOT.} 
Most 3D MOT methods~\cite{immortal_tracker, simpletrack} still follow the tracking-by-detection paradigm, which benefits from off-the-shelf SOTA 3D detectors.
AB3DMOT~\cite{ab3dmot} associates the detected box with tracked trajectories by Kalman filter~\cite{kalman} and Hungarian algorithm.
CenterPoint~\cite{centerpoint} calculates the previous position of the detected box with predicted velocity to get similarity with tracked box and solves matching pair by a greedy algorithm.
Pre-processing detection boxes and GIoU-based two-stage data association strategy show effectiveness to improve the performance~\cite{simpletrack}. Besides, the mismatches are sharply decreased by enlarging the maximum death age to alleviate early termination~\cite{immortal_tracker} on WOD.

\noindent \textbf{Sequential Point Clouds Learning.} 
With the emergence of large-scale LiDAR serialized point cloud datasets~\cite{sun2020scalability,caesar2020nuscenes,geiger2013vision,behley2019semantickitti}, researchers are increasingly exploring the use of sequential point clouds in real-world scenarios, such as multi-frame object detection~\cite{mppnet, yang20213d, int}, point cloud segmentation~\cite{aygun20214d, marcuzzi2022contrastive, hurtado2020mopt}, object tracking~\cite{ab3dmot, simpletrack, immortal_tracker}, and scene flow prediction~\cite{liu2019flownet3d, gu2019hplflownet, wu2020motionnet}.
Notable works in this area include 3DAL~\cite{3dal} and Auto4D~\cite{auto4d}, which refine detection boxes at the trajectory level with human input or off-the-shelf detectors. 
Both of them contain some components that follow the sliding window fashion, which ignores the importance of the long-term characteristics of the tracks.
These works inspired our own research efforts in this area.

\section{Methodology}
We first give a brief review of the entire pipeline of offboard 3D object detection. Meanwhile, the core problem is stated by introducing the input and output representations in Sec.~\ref{sec:prelim}. Afterward, we present how to boost the potential of the overall pipeline by improving both the upstream object tracks generation in Sec.~\ref{sec:upstream}, and the downstream attribute-based refining in Sec.~\ref{sec:downstream}. Details of the network architecture, losses and training strategy are described in Appendix.

\begin{figure*}[t]
\centering
\includegraphics[width=1.\textwidth]{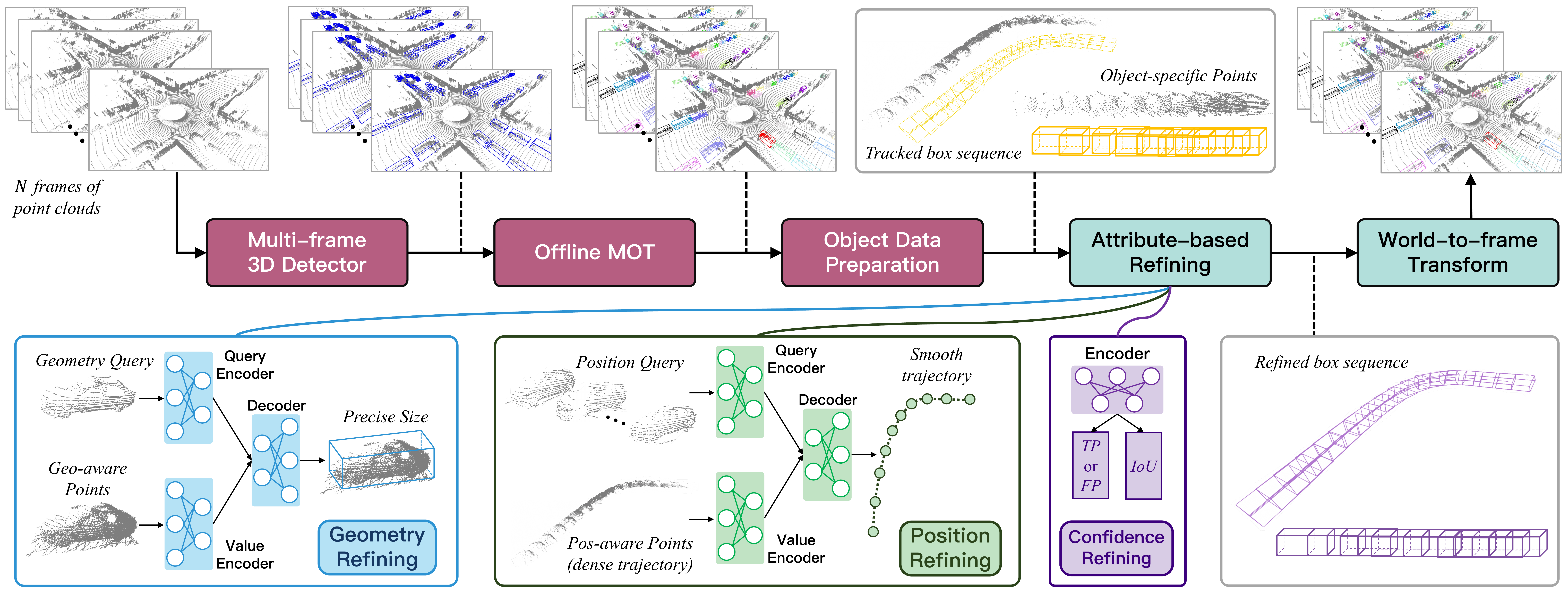}
\caption{Overview of the framework of our proposed DetZero. The multi-frame detector takes as input $N$ frames of point clouds, the following offline tracker generates accurate and complete object tracks. For each object track, we prepare its object-specific LiDAR points sequence and tracked box sequence. Consequently, we refine the object tracks through three simultaneous steps: refine the geometry size, smooth the motion trajectory and update the confidence score. Afterwards, they are combined together and transformed through world-to-frame poses as the final ``auto labels''.}
\label{fig:framework}
\end{figure*}

\subsection{Preliminary}
\label{sec:prelim}
We select 3DAL~\cite{3dal} as our baseline, which is a SOTA offboard 3D detector consisting of four modules to process a sequence of point clouds.
Specifically, the first detection module takes as input $N$ frames of point clouds $\left\{ \mathcal{P}_i \in \mathbb{R}^{n_i \times (3+C)} \ | \ i=1,2,...,N \right\}$ ($n_i$ points for each frame, $C$ is the additional feature for each point such as intensity and elongation) and outputs frame-level 3D bounding boxes $\hat{b}_i \in \mathbb{R} ^ {m_i \times 7} $ and categories. Then, a multi-object baseline tracker links the detected objects across frames as continuous object tracks $\{ T_{j}\in\mathbb{R}^{L_{j}\times 7}, j \in N_{obj} \}$ ($L$ is track length) with the unique object IDs. For each object track $T_{j, i}$ ($j$-th object at frame $i$), the object-specific LiDAR points are extracted by cropping original point clouds within corresponding bounding boxes, which are then merged together by eliminating ego-motion with frame poses $\left\{ \mathcal{M}_i = [ \mathbf{R}_i|\mathbf{t}_i ] \in \mathbb{R}^{3 \times 4} \right\}$.
The third motion classification module is utilized to determine an object’s motion state (static or dynamic) based on its trajectory features. In the final step, the object-centric auto labeling models extract the object's temporal representation separately based on its predicted motion state, to predict precise boxes. The refined boxes are eventually transferred back to each frame with the inverse frame pose. Please refer to the original paper for more details~\cite{3dal}.

There are two factors that affect object-specific temporal context learning: \textit{incomplete tracks} from upstream module and \textit{motion-state-based} auto labeling models that ignore common object characteristics. Incomplete object tracks hinder the generation of effective object-specific temporal point cloud data, illustrated in Fig.~\ref{fig:track-prob}. The sliding-window-based dynamic object refining mechanism fails to use complete temporal contexts, such as the relation between local position and global trajectory and object geometry consistency, as depicted in Fig.~\ref{fig:refiner-prob}.

These observations prompt reconsideration of current offboard 3D object detection conventions. As is illustrated in Fig.~\ref{fig:framework}, our evolution focuses on two main aspects: (1) using a multi-frame 3D detector and offline tracker to provide sufficient accurate and complete object tracks, and (2) modernizing the attention-mechanism refining module to reason about object attribute representations in long-term sequential point clouds. Both aspects significantly impact the model's performance, yet were not thoroughly investigated in prior studies.

\subsection{Complete Object Tracks Generation}
\label{sec:upstream}
Our upstream object detection and tracking module aims to generate accurate and complete object tracks, which is essential as the entry point to the whole pipeline.

\noindent \textbf{Object Detection.}
The competitive CenterPoint~\cite{centerpoint} is adopted as our base detector because the anchor-free design would predict dense and redundant bounding boxes. To provide accurate prediction results as much as possible, we strengthen it in three aspects: (1) a combination of five frames of point cloud serves as the input to maximize the contributions rather than performance diminishes~\cite{3dal,mppnet}; (2) a point density aware module is designed to leverage the raw point features and voxel feature for precise refinement~\cite{pdv}; (3) to improve the adaption towards complex surroundings, test time augmentation~\cite{tta} (TTA) for point cloud data and multi-model ensemble (different resolution, network structure and capacity) are utilized to boost the detection performance. Please see the details in Appendix.

\noindent \textbf{Offline Tracking.}
Recent multi-object trackers~\cite{ab3dmot}, taking the tracking-by-detection path, always struggle in redundant detected bounding boxes which focus much on box-level detection metrics.
Taking inspiration from~\cite{simpletrack,immortal_tracker}, our multi-object tracker utilizes a two-stage data association strategy to mitigate the possibility of false matching. Concretely, the detected boxes are partitioned into two distinct groups based on their confidence scores. The pre-existing object tracks initially engage in data association solely with the high-score group, and subsequently, successfully associated boxes are utilized to update the existing tracks. The un-updated tracks are further associated with the low-score group, and the un-associated boxes are deprecated. In addition, the life cycle of an object is allowed to persist immortally until the sequence terminates, after which any redundant boxes that have not been updated are removed. This operation benefits the re-connection of truncated object tracks and effectively prevents ID switches.

Post-processing is also crucial to generate good-quality tracks. We re-execute our tracking method following reverse time order to generate another group of tracks $T_{i,j}^\text{inv}$. These tracks are then associated together by a location-aware similarity matching score. Finally, we fuse the paired tracks with the WBF~\cite{wbf} strategy to further ameliorate the missing boxes and stable the motion state, which is called \textit{forward and reverse order tracking fusion}. Besides, we do not run the downstream modules for too short tracks. The boxes of those short tracks, together with the redundant boxes that have not been updated are merged directly into the final auto labels.

\noindent \textbf{Object Data Preparation.}
Given an object track (identified by the unique object ID), we first slightly scale up the RoI area of the tracked boxes by a parameter $\alpha$ along three dimensions, which compensates for abundant contextual information. Then, points that lie within the regions bounded by these enlarged boxes are taken out. We denote this object-specific LiDAR points sequence as $\{ \mathcal{P}_{j, i}\}$ for object $j$ with length $L_j$ at $i$-th frame of the original sequence, as well as its corresponding tracked box sequence $\{ \hat{b}_{j, i} \}$ and confidence scores $\{ S_{j, i} \}$.

\subsection{Attribute-based Refining Module}
\label{sec:downstream}
Previous object-centric auto labeling methods have employed a state-based strategy to refine the proposals generated by upstream modules. This approach not only results in the propagation of misclassification but also disregards the potential similarities between objects. However, it has been observed that, for rigid objects, the geometric shape of an object does not vary significantly over a continuous period of time, regardless of its motion state. Furthermore, an object's motion state typically exhibits regular patterns and strong consistency with neighboring moments. Based on these observations, we propose a novel approach that decomposes the traditional bounding box regression task into three distinct modules that predict an object's geometry, position, and confidence attributes, respectively.

\subsubsection{Geometry Refining Model}

\noindent \textbf{Geometry-aware Points Generation.}
The acquisition of complementary information regarding the appearance and shape of an object can be facilitated by obtaining multiple viewpoints of an object. To obtain a cohesive object track, a local coordinate transform operation, involving translation and rotation, similar to the method presented in~\cite{pointrcnn,frustum}, is initially applied to align the object points to a local box coordinate at various locations. Subsequently, points from different frames are amalgamated, irrespective of their original origin.
Henceforth, we randomly sample a set of points $(\mathcal{P}_j = \{ p_1, ..., p_n \} \in \mathbb{R}^{n \times (3+C)}, n = 4096)$ for further processing.

\noindent \textbf{Proposal-to-Point Encoding.}
It is of paramount importance to effectively utilize the geometric information by encoding proposals into object points rather than discarding them after object points extraction~\cite{ct3d}.
Specifically, for each point $p_k$ of $\mathcal{P}_j$ and its corresponding box $\hat{b}_{j,i}$, we use a point-to-surface approach to compute the projection distance between $p_k$ and the six surfaces of $\hat{b}_{j,i}$, denoted as $\Delta p_k^\text{sf}$.
The newly generated point features can be viewed as a better representation of proposal information, which can be expressed as $\left[p_k, \Delta p_k^{\text{sf}_1}, \cdots, \Delta p_k^{\text{sf}_6} \right]$, where $[\cdot]$ denotes the concatenate operation.

\noindent \textbf{Attention-based Geometry Interaction across Views.}
It has been investigated in 3D object detection that a better initialization of object queries would benefit the convergency of the transformer network~\cite{transfusion}. Inspired by this observation, we propose to initialize the geometry query features based on object-specific points.
Firstly, we randomly select $t$ samples from the whole object track. Each sample has corresponding $256$ randomly-selected points. Each point is also augmented by our proposed proposal-to-point encoding approach, and besides, the corresponding confidence scores.
Afterwards, a PointNet-structure encoder $\text{ENC}_1$ is adopted to extract features for each selected sample, which is used to initialize as the geometry queries $\mathbf{Q}^\text{geo} \in \mathbb{R}^{t \times D} $.
Then we utilize another encoder $\text{ENC}_2$ to take as input $\mathcal{P}_j$ and extract dense point features, which are served as $\mathbf{K}^\text{geo}$ and $\mathbf{V}^\text{geo} \in \mathbb{R}^{n \times D} $.

The generated geometry queries are first fed into the multi-head self-attention layer, to encode rich contextual relationships among selected samples and feature dependencies for refining geometry information.
The following cross attention between geometry queries and the point features aggregates relevant context onto the object candidates, which reasons pairwise differences to compensate the point features of supplementary views for each geometry query.
At last, a feed-forward network (FFN) independently decodes $t$ geometry queries into $t$ geometry sizes, which are then averaged as the final predicted size.

For better residual target regression, we map the proposals' size to $D$-dim embeddings with a linear projection layer. They are element-wisely summed with the query features.
Details of the network architecture are shown in Appendix.

\subsubsection{Position Refining Model}

\noindent \textbf{Position-aware Points Generation.}
For $j$-th object, we randomly select the position of a box from its tracked box sequence $\{ \hat{b}_{j,i} \}$ as a new local coordinate system, and subsequently, the other boxes are transformed to this coordinate, as well as the corresponding object-specific points $\{ \mathcal{P}_{j, i} \}$. Then, a fixed number of points are randomly selected from $\{ \mathcal{P}_{j, i} \}$ for each frame.

For each point, in addition to calculating the distance to the proposal's center, we also compute the relative coordinates between each point and eight corners of the corresponding tracked box as $\Delta p_{k}^\text{co} = p_k - p^\text{co}$, which results in a $27$-dim feature vector. The final position-aware point features can be expressed as $ f_k^\text{pos} =  \left[ p_k, \Delta p_k^\text{ce}, \Delta p_k^{\text{co}_1}, \cdots, p_k^{\text{co}_8} \right]$.
To facilitate training, all object tracks are padded to the same length with zeros.

\noindent \textbf{Attention-based Local-to-Global Position Interaction.}
For an object track, we utilize the same structured query encoder as $\text{ENC}_1$ in Geometry Refining Model (GRM) to generate position queries $\mathbf{Q}^\text{pos} \in \mathbb{R}^{L \times D}$ for $L$ frame, whose features consist of position-aware features $f^\text{pos}$ and confidence scores. Simultaneously, we extract the point features of the entire object track using another encoder that takes $f^\text{pos}$ as input. These features serve as $\mathbf{K}^\text{pos}$ and $\mathbf{V}^\text{pos} \in \mathbb{R}^{n^\text{pos} \times D}$ for subsequent computation, where $n^{pos}$ is the number of sampled points.

The position queries are first fed into the self-attention module, to capture the relative distance between itself and others. Additionally, we apply a 1D mask near the position of each query to weigh the self-attention.
Subsequently, the local position queries $\mathbf{Q}^\text{pos}$ and global point trajectory features $\mathbf{K}^\text{pos}$, $\mathbf{V}^\text{pos}$ are fed into the cross-attention module to model the local-to-global position contextual relations.
Finally, we predict the offsets between each ground-truth center and the corresponding initial center under the local coordinate system, as well as the bin-based heading angle.

\begin{table*}[]
\footnotesize
\setlength{\tabcolsep}{0.09cm}
\renewcommand\arraystretch{1.1}

  \begin{center}
  \begin{tabular}{l|c|c|c|cc|cc|cc}
    \Xhline{0.75pt}
    \multirow{2}{*}{ Method } & \multirow{2}{*}{Rank} & \multirow{2}{*}{Frames} & mAPH & \multicolumn{2}{c|}{\textit{Vehicle} (AP/APH)} & \multicolumn{2}{c|}{\textit{Pedestrian} (AP/APH)} & \multicolumn{2}{c}{\textit{Cyclist} (AP/APH)} \\

    {} & {} & {} & L2 & L1 & L2 & L1 & L2 & L1 & L2 \\
    \hline
    $\text{DetZero\_Ens}^{\dagger}$ (Ours) & $1$ & $200$ & $\mathbf{85.15}$ & $\mathbf{92.17}$ / $\mathbf{91.80}$ & $\mathbf{87.32}$ / $\mathbf{86.95}$ & $\mathbf{92.08}$ / $\mathbf{89.67}$ & $\mathbf{88.08}$ / $\mathbf{85.64}$ & $85.49$ / $84.70$ & $83.63$ / $82.85$ \\

    $\text{3DAL}^{\dagger}$~\cite{3dal} & $-$ & $200$ & $-$ & $85.84$ / $85.46$ & $77.24$ / $76.91$ & $-$ & $-$ & $-$ & $-$ \\
    
    \hline
    $\text{LoGoNet\_Ens}^{\ddagger}$~\cite{logonet} & $2$ & $5$ & $81.96$ & $88.80$ / $88.37$ & $82.75$ / $82.32$ & $89.63$ / $86.74$ & $84.96$ / $82.10$ & $84.51$ / $83.59$ & $82.36$ / $81.46$ \\

    $\text{HRI\_ADLAB\_HZ}^{*}$ & $3$ & $4$ & $81.32$ & $86.77$ / $86.40$ & $80.19$ / $79.83$ & $88.59$ / $86.01$ & $83.84$ / $81.27$ & $\mathbf{85.67}$ / $\mathbf{84.84}$ & $\mathbf{83.69}$ / $\mathbf{82.87}$ \\

    $\text{MT-Net v2}$\cite{mt-net} & $4$ & $5$ & $80.00$ & $87.54$ / $87.12$ & $81.20$ / $80.79$ & $87.62$ / $84.89$ & $82.33$ / $79.66$ & $82.80$ / $81.74$ & $80.58$ / $79.54$ \\

    $\text{BEVFusion\_TTA}^{\ddagger}$~\cite{bevfusion} & $5$ & $3$ & $79.97$ & $87.96$ / $87.58$ & $81.29$ / $80.92$ & $87.64$ / $85.04$ & $82.19$ / $79.65$ & $82.53$ / $81.67$ & $80.17$ / $79.33$ \\

    $\text{LidarMultiNet\_TTA}$~\cite{lidarmultinet} & $6$ & $3$ & $79.94$ & $87.64$ / $87.26$ & $80.73$ / $80.36$ & $87.75$ / $85.07$ & $82.48$ / $79.86$ & $82.77$ / $81.84$ & $80.50$ / $79.59$ \\

    $\text{MPPNet\_TTA}$~\cite{mppnet} & $7$ & $16$ & $79.60$ & $87.77$ / $87.37$ & $81.33$ / $80.93$ & $87.92$ / $85.15$ & $82.86$ / $80.14$ & $80.74$ / $79.90$ & $78.54$ / $77.73$ \\

    $\text{LIVOX Detection}^{*}$ & $9$ & $7$ & $78.96$ & $86.81$ / $86.45$ & $80.41$ / $80.05$ & $87.17$ / $84.57$ & $82.16$ / $79.59$ & $80.46$ / $79.59$ & $78.08$ / $77.24$ \\
    $\text{DeepFusion\_Ens}^{\ddagger}$~\cite{deepfusion} & $12$ & $5$ & $78.41$ & $86.45$ / $86.09$ & $79.43$ / $79.09$ & $86.14$ / $83.77$ & $80.88$ / $78.57$ & $80.53$ / $79.80$ & $78.29$ / $77.58$ \\

    $\text{AFDetV2-Ens}$~\cite{afdetv2} & $17$ & $1$ & $77.64$ & $85.80$ / $85.41$ & $78.41$ / $78.34$ & $85.22$ / $82.16$ & $79.71$ / $76.75$ & $81.20$ / $80.30$ & $78.70$ / $77.83$ \\
    $\text{INT\_Ens}$~\cite{int} & $19$ & $100$ & $77.21$ & $85.63$ / $85.23$ & $79.12$ / $78.73$ & $84.97$ / $81.87$ & $79.35$ / $76.36$ & $79.76$ / $78.65$ & $77.62$ / $76.54$ \\

    \Xhline{0.75pt}
  \end{tabular}
  \end{center}
  \caption{Performance comparison on the Waymo 3D detection leaderboard. Metrics are standard 3D AP and APH by both L1 and L2 difficulties. Note that all the listed entries use TTA or model ensemble techniques.
  We use $\dagger$ to denote the offboard 3D detectors. LiDAR-Camera fusion based 3D detectors are marked with $\ddagger$, and anonymous submissions are marked with $*$. We report the performance till 2023-03-08 23:59 GMT.}
  \label{table:leaderboard}
\end{table*}

\subsubsection{Confidence Refining Model}
Our detection and offline tracking module is encouraged to generate sufficient object tracks, which naturally contain boxes that are far from being true positive, even after GRM and Position Refining Model (PRM). To address this issue, we introduce a Confidence Refining Model (CRM), composed of two branches to optimize the confidence scores.

The first classification branch is similar to the traditional second-stage object detector~\cite{fasterrcnn}, for determining TPs or FPs by updating scores. 
We assign negative labels to the tracked boxes whose IoU ratios with corresponding ground-truth boxes are lower than $\tau_{l}$.
Tracked boxes with IoU ratio higher than $\tau_{h}$ are treated as positive samples. Other boxes do not contribute to the classification objective.

The second IoU regression branch predicts how much IoU an object should have after being refined~\cite{ciassd}. Hence, the regression targets are set as the IoUs between ground-truth boxes and refined ones predicted by previous GRM and PRM.

In the beginning, we process the object-specific points with a similar network structure encoder identical to ENC$_1$ of GRM. The extracted point cloud features are fused by a simple MLP, and then fed into these two branches for predicting respective scores.
During training, we randomly sample pre-divided positive and negative object tracks with $1:1$ ratio in each epoch for better convergence. The final scores are the geometric average of the two branches: $S_j = \sqrt{S_{j,\text{cls}}^{2}+S_{j,\text{iou}}^{2}}$.

\section{Experiments}
In this section, we first introduce the dataset details and evaluation metrics used in our experiments. We then provide a detailed performance comparison between our DetZero and other SOTA 3D detectors in Sec.~\ref{sec:result-sota}. Then, we validate whether such high-quality ``auto labels'' by DetZero could play the same role as human labels in Sec.~\ref{sec:result-human}. In Sec.~\ref{sec:ablation}, we present the ablation studies and analysis for convincing each component of our entire approach. Please refer to Appendix for more detailed experiments and ablation results.

\subsection{Dataset}
We conduct experiments on the challenging Waymo Open Dataset~\cite{wod}, which is one of the largest dataset containing total 1150 LiDAR scenes with 798 for training, 202 for validation and 150 for testing. The dataset provides 20-second point clouds data for each scene with a sampling frequency at 10Hz, and 3D annotations for 4 object categories in 360 degree field of view. We follow the evaluation protocol with the official metrics, \textit{i.e.}, average precision (AP) and average precision weighted by heading (APH), and report the results on both LEVEL 1 (L1) and LEVEL 2 (L2) difficulty levels. The L1 evaluation includes objects with more than five LiDAR points and L2 evaluation only includes 3D labels with at least one and no more than five LiDAR point. Note that mAPH (L2) is the main metric for ranking in the Waymo 3D detection challenge.

\begin{table*}[]
\footnotesize
\setlength{\tabcolsep}{0.55cm}
\renewcommand\arraystretch{1.1}

  \begin{center}
  \begin{tabular}{l|c|cc|cc}
    \Xhline{0.75pt}
    \multirow{2}{*}{ Method} & \multirow{2}{*}{Frames} & \multicolumn{2}{c|}{\textit{Vehicle} (AP/APH)} & \multicolumn{2}{c}{\textit{Pedestrian} (AP/APH)} \\
     &  & L1 & L2 & L1 & L2 \\
    
    \hline
    $\text{MVF}$~\cite{mvf} & $1$ & $62.93$ / $-$ & $-$ / $-$ & $65.33$ / $-$ & $-$ / $-$ \\

    $\text{PV-RCNN}$~\cite{pvrcnn} & $1$ & $77.51$ / $76.89$ & $68.98$ / $68.41$ & $75.01$ / $65.65$ & $66.04$ / $57.61$ \\
    $\text{CenterPoint}$~\cite{centerpoint} & $1$ & $76.7$ / $76.2$ & $68.8$ / $68.3$ & $79.0$ / $72.9$ & $71.0$ / $65.3$ \\

    $\text{PDV}$~\cite{pdv} & $1$ & $76.85$ / $76.33$ & $69.30$ / $68.81$ & $74.19$ / $65.96$ & $65.85$ / $58.28$ \\

    $\text{INT}$~\cite{int} & $2$ & $-$ / $-$ & $-$ / $73.3$ & $-$ / $-$ & $-$ / $71.9$ \\

    $\text{3D-MAN}$~\cite{3dman} & $16$ & $74.50$ / $74.00$ & $67.60$ / $67.10$ & $71.10$ / $67.70$ & $62.60$ / $59.00$ \\
    $\text{CenterFormer}$~\cite{centerformer} & $8$ & $78.80$ / $78.30$ & $74.30$ / $73.80$ & $82.10$ / $79.30$ & $77.80$ / $75.00$ \\
    
    $\text{3DAL}^{\dagger}$~\cite{3dal} & $200$ & $84.50$ / $-$ & $-$ / $-$ & $82.88$ / $-$ & $-$ / $-$ \\

    $\text{MPPNet}$~\cite{mppnet} & $16$ & $82.74$ / $82.28$ & $75.41$ / $74.96$ & $84.69$ / $82.25$ & $77.43$ / $75.06$ \\

    \hline
    $\text{DetZero}^{\dagger}$ (Upstream) & 5 & ${83.07}$ / ${82.57}$ & ${75.72}$ / ${75.24}$ & ${86.17}$ / ${83.07}$ & ${79.39}$ / ${76.34}$ \\
    
    $\text{DetZero}^{\dagger}$ (Full) & $200$ & $\mathbf{89.49}$ / $\mathbf{89.06}$ & $\mathbf{83.34}$ / $\mathbf{82.92}$ & $\mathbf{89.54}$ / $\mathbf{87.06}$ & $\mathbf{83.52}$ / $\mathbf{81.01}$ \\

    \Xhline{0.75pt}
  \end{tabular}
  \end{center}
  \caption{Performance comparison on the val set of WOD. Metrics are standard 3D AP and APH by both L1 and L2 difficulties. We use $\dagger$ to denote the entries using TTA or model ensemble techniques.}
  \label{table:valcompare}
\end{table*}

\begin{table}
\small
\renewcommand\arraystretch{1.1}

  \begin{center}
  \begin{tabular}{c|cc|cc}
    \Xhline{0.75pt}
    & \multicolumn{2}{c|}{3D AP} & \multicolumn{2}{c}{BEV AP} \\
    
    & {IoU=0.7} & {IoU=0.8} & {IoU=0.7} & {IoU=0.8} \\
    
    \hline
    {Human} & {$86.45$} & {$60.49$} & {$93.86$} & {$86.27$} \\
    
    {3DAL} & {$85.37$} & {$56.93$} & {$92.80$} & {$87.55$} \\
    
    {DetZero (Ours)} & {$\mathbf{90.24}$} & {$\mathbf{67.61}$} & {$\mathbf{95.15}$} & {$\mathbf{90.04}$} \\
    
    \Xhline{0.75pt}
  \end{tabular}
  \end{center}
  \caption{Comparing human labels and auto labels. The results are 3D and BEV AP (L1 difficulty) under 0.7 and 0.8 IoU threshold for \textit{Vehicle} on 5 sequences selected from WOD val set. Please refer to the appendix for more details about the sequences' IDs and human's AP computing method.}
  \label{table:human}
\end{table}

\subsection{Comparising with State-of-the-art Detectors}
\label{sec:result-sota}
We present a comprehensive comparison of our DetZero with various state-of-the-art 3D detectors.

As shown in Table~\ref{table:leaderboard}, our DetZero achieves the best results on Waymo 3D detection challenge leaderboard~\cite{wod_3d_detection_website} with $85.15$ mAPH (L2) detection performance.
For comparisons among methods processing long-term sequential point clouds (at least 100 frames), DetZero surpasses 3DAL~\cite{3dal} with $5.93$ (L1) and $9.51$ (L2) mAPH on \textit{Vehicle}, surpasses INT~\cite{int} with $6.16$ (L1) and $7.69$ (L2) mAPH on \textit{Vehicle}, and $7.65$ (L1) and $9.09$ (L2) mAPH on \textit{Pedestrian}. DetZero shows great ability to leverage the long-term sequential point clouds for offboard perception.
Moreover, compared to state-of-the-art multi-modal fusion 3D detectors~\cite{logonet,bevfusion,deepfusion}, DetZero also yields a strong performance gain with at least $3.43$ (L1) and $4.63$ (L2) mAPH on \textit{Vehicle}, and $2.93$ (L1) and $3.54$ (L2) mAPH on \textit{Pedestrian}. These results further highlight the great potential of the point cloud sequences explored by DetZero.
We also ranked $1$st place on Waymo 3D tracking challenge leadboard~\cite{wod_3d_tracking_website} with $75.05$ MOTA (L2) by a $9.97$ point margin, please see the detailed performance in Appendix.

Additionally, in Table~\ref{table:valcompare}, we provide a comparison between SOTA 3D detectors and our internal components on the validation set of WOD.
We outperform other single-frame and multi-frame based methods with a huge margin on both \textit{Vehicle} and \textit{Pedestrian}. Thanks to the high-quality object tracks generated by our upstream module, our full model gets a significant internal improvement: $6.49$ (L1) and $7.68$ (L2) mAPH for \textit{Vehicle}, $3.99$ (L1) and $4.67$ (L2) mAPH for \textit{Pedestrian}, more analysis is shown in Sec.~\ref{sec:ablation}.

\subsection{Comparising with Human Labels}
\label{sec:result-human}
It has been shown that humans' capability of recognizing objects in a dynamic 3D scene has minor fluctuations~\cite{3dal}. Human performance is measured by the consistency between the humans’ single-frame-based re-labeling and released multi-frame-based ground-truth labels.

We follow their experimental setup to report the mean AP of our DetZero across the 5 selected sequences.
In Table~\ref{table:human}, we demonstrate superior performance compared to human and 3DAL in particular. With the common 3D AP@0.7 metric, we achieve $3.79$ and $4.87$ points gains, while the gap is larger in more strict 3D AP@0.8 metric. We obtain similar gains with the BEV AP by ignoring height. To the best of our knowledge, this is the first time that the offboard 3D detector model can outperform the average human labels.

\begin{table}
  \begin{center}
  \small
  \setlength{\tabcolsep}{0.16cm}
  \renewcommand\arraystretch{1.1}
  \begin{tabular}{c|cc|cc}
    \Xhline{0.75pt}
    \multirow{2}{*}{Training Data} & \multicolumn{2}{c|}{\textit{Vehicle}} & \multicolumn{2}{c}{\textit{Pedestrian}} \\
    
     & {AP} & {APH} & {AP} & {APH} \\
    
    \hline
    {100\% \textit{train}  (Human)} & {$\mathbf{75.41}$} & {$\mathbf{74.88}$} & {$77.51$} & {$71.16$} \\

    \hdashline
    {10\% \textit{train}(Human)} & {$66.88$} & {$66.28$} & {$67.13$} & {$59.66$} \\

    \hline
    {90\% \textit{train} (DetZero)} & {$74.12$} & {$73.59$} & {$78.57$} & {$71.39$} \\

    \hdashline
    
    10\% \textit{train} (Human) & \multirow{2}{*}{$74.44$} & \multirow{2}{*}{$73.91$} & \multirow{2}{*}{$\mathbf{78.92}$} & \multirow{2}{*}{$\mathbf{72.02}$} \\
    + 90\% \textit{train} (DetZero) & & & \\
    
    \Xhline{0.75pt}
  \end{tabular}
  \end{center}
  \caption{Intra-domain semi-supervised learning results.}
  \label{table:intra-domain}
\end{table}

To better study whether such high-quality auto labels could replace human labels for onboard model training, we conduct another intra-domain semi-supervised learning experiment. We choose the single-stage CenterPoint~\cite{centerpoint} as our student model. Note that the student model takes as input a single frame, and the GT-Paste data augmentation is not used during training.
We first randomly select 10\% sequences (79 ones) in the WOD training set to train our entire DetZero pipeline. Next, we can generate ``auto labels'' for the rest 90\% sequences (719 ones) in the training set. Afterwards, the student model is trained with different combinations of human labels and ``auto labels''.

\begin{table*}
\small
\setlength{\tabcolsep}{0.38cm}
\renewcommand\arraystretch{1.1}
  \begin{center}
  \begin{tabular}{ccccc|cc|cc}
    \Xhline{0.75pt}
    \multirow{2}{*}{Det.} & \multirow{2}{*}{Tra.} & \multirow{2}{*}{GRM} & \multirow{2}{*}{PRM} & \multirow{2}{*}{CRM} & \multicolumn{2}{c|}{\textit{Vehicle} (L1 / L2)} & \multicolumn{2}{c}{\textit{Pedestrian} (L1 / L2)} \\
     &  &  &  &  & {IoU=0.7} & {IoU=0.8} & {IoU=0.5} & {IoU=0.6} \\ 
    \hline
    {\checkmark} & {} & {} & {} & {} & {$82.57$ / $75.09$} & {$51.34$ / $44.77$} & {$83.23$ / $76.47$} & {$64.12$ / $56.49$} \\

    {\checkmark} & {\checkmark} & {} & {} & {} & {$82.57$ / $75.24$} & {$51.34$ / $44.81$} & {$83.07$ / $76.34$} & {$64.04$ / $56.44$} \\

    {\checkmark} & {\checkmark} & {\checkmark} & {} & {} & {$84.49$ / $77.17$} & {$56.71$ / $49.60$} & {$84.71$ / $78.04$} & {$68.33$ / $60.48$} \\

    {\checkmark} & {\checkmark} & {} & {\checkmark} & {} & {$85.48$ / $78.55$} & {$56.35$ / $49.56$} & {$84.32$ / $77.78$} & {$66.53$ / $58.99$} \\

    {\checkmark} & {\checkmark} & {\checkmark} & {\checkmark} & {} & {$87.81$ / $81.01$} & {$64.53$ / $57.15$} & {$85.94$ / $79.48$} & {$70.97$ / $63.26$}\\

    {\checkmark} & {\checkmark} & {\checkmark} & {\checkmark} & {\checkmark} & {$\mathbf{89.06}$ / $\mathbf{82.92}$} & {$\mathbf{64.94}$ / $\mathbf{57.84}$} & {$\mathbf{87.06}$ / $\mathbf{81.01}$} & {$\mathbf{71.61}$ / $\mathbf{64.03}$} \\
    
    \Xhline{0.75pt}
  \end{tabular}
  \end{center}
  \caption{Effect of each component in our DetZero on WOD val set. Metrics are 3D APH of both L1 and L2 difficulties for \textit{Vehicle} and \textit{Pedestrian} with a standard IoU threshold (0.7 \& 0.5) and a higher IoU threshold (0.8 \& 0.6).}
  \label{table:ablations}
\end{table*}

As shown in Table~\ref{table:intra-domain}, the first two rows give a performance comparison by reducing the human annotations to 10\%, which decreases the student model's performance by $8.53$ and $8.6$ points for \textit{Vehicle}, $10.38$ and $11.5$ points for \textit{Pedestrian}.
Surprisingly, when we add other 90\% auto labels, the performance increase with $7.56$ and $7.63$ points for \textit{Vehile} which is close to the first row, $11.79$ and $12.36$ points for \textit{Pedestrian} which is already higher. Besides, when we remove the 10\% human labels (3rd row), the results are predictably slightly lower than the 4th row, still showing $1.06$ and $0.23$ gain for \textit{Pedestrian}.
These results demonstrate that ``auto labels'' generated by our DetZero are qualified for training online models.
We visualized the results and found that ``auto labels'' contain fewer pedestrian labels than human labels, such as the hard samples at a far distance. Hence, the student model trained with our ``auto labels'' would output fewer false positives compared to the model trained with 100\% human labels. More detailed analyses are in Appendix.

\vspace{6mm}
\subsection{Ablation Studies and Analysis}
\label{sec:ablation}
We conduct ablation studies on the WOD validation set to verify all the components of our approach, especially under a fair experimental setting regardless of the techniques for the leaderboard. Additional ablations for the network structures and data augmentations are shown in Appendix.

\noindent \textbf{Effects of each Component.}
We enable different combinations of our proposed modules to evaluate the performances. In addition to the commonly used standard IoU threshold, we also report the performance under a higher IoU threshold to more accurately assess the disparity between predictions and the ground-truth labels.

In Table~\ref{table:ablations}, compared to the upstream results (2nd row), when IoU equals $0.7$, the 3rd row shows that the GRM gains $1.92$ (L1) and $1.93$ (L2) points for \textit{Vehicle}, and $1.64$ and $1.7$ points for \textit{Pedestrian}. As a comparison, the 4th row shows that the PRM gains $2.91$, $3.31$, $1.25$ and $1.44$ points respectively. When we combine them together, the performance improves a lot, shown by the 5th row. And the CRM also performs well, by re-scoring the samples based on their qualities.
When IoU equals $0.8$, we get impressive improvements. Specifically, our full downstream refining module boosts the performance by $26.49\%$ and $29.08\%$ for \textit{Vehicle}, by $11.82\%$ and $13.45\%$ for \textit{Pedestrian}. This shows that our entire DetZero tries its best to generate high-quality 3D boxes.

\begin{table}
\small
\setlength{\tabcolsep}{0.15cm}
\renewcommand\arraystretch{1.1}
  \begin{center}
  \begin{tabular}{cccc|cc|cc}
    \Xhline{0.75pt}
      \multirow{2}{*}{Trk$_1$} & \multirow{2}{*}{Trk$_2$} & \multirow{2}{*}{Ref$_1$} & \multirow{2}{*}{Ref$_2$} & \multicolumn{2}{c|}{\textit{Vehicle}} & \multicolumn{2}{c}{\textit{Pedestrian}} \\
    
     &  &  &  & {L1} & {L2} & {L1} & {L2} \\
    
    \hline
    {\checkmark} & {} & {\checkmark} & {}  & {$83.74$} & {$76.33$} & {$83.92$} & {$76.94$} \\

    {\checkmark} & {} & {} & {\checkmark} & {$85.57$} & {$78.14$} & {$85.47$} & {$78.24$} \\

    {} & {\checkmark} & {\checkmark} & {} & {$84.92$} & {$77.06$} & {$84.64$} & {$77.52$} \\
    
    {} & {\checkmark} & {} & {\checkmark} & 
    {$\mathbf{89.06}$} & {$\mathbf{82.92}$} & {$\mathbf{87.06}$} & {$\mathbf{81.01}$} \\
    
    \Xhline{0.75pt}
  \end{tabular}
  \end{center}
  \caption{Evaluate the function of different upstream and downstream modules. We reproduce 3DAL~\cite{3dal} and use the subscripts $1$ and $2$ to represent their and our models respectively. Metrics are standard 3D APH of both L1 and L2 difficulties for \textit{Vehicle} and \textit{Pedestrian}.
  }
  \label{table:cross-eval}
\end{table}

\begin{table}
\small
\setlength{\tabcolsep}{0.15cm}
\renewcommand\arraystretch{1.1}
  \begin{center}
  \begin{tabular}{ccc|cc|cc}
    \Xhline{0.75pt}
      \multirow{2}{*}{Trk$_1$} & \multirow{2}{*}{Trk$_2$} & \multirow{2}{*}{Ref$_2$} & \multicolumn{2}{c|}{MOTA} & \multicolumn{2}{c}{Recall@track} \\

     & & & {\textit{Vehicle}} & {\textit{Pedestrian}} & {\textit{Vehicle}} &{\textit{Pedestrian}} \\
    
    \hline
    {\checkmark} & {} & {} & {$57.14$}  & {$61.78$} & {$23.96$} & {$26.63$} \\

    {} & {\checkmark} & {} & {$58.41$}  & {$62.50$} & {$40.28$} & {$45.19$} \\

    {\checkmark} & {} & {\checkmark} & {$63.24$} & {$65.52$}   & {$27.41$} & {$27.77$} \\
    
    {} & {\checkmark} & {\checkmark} & 
    {$\mathbf{71.36}$}  & {$\mathbf{68.78}$} & {$\mathbf{57.79}$} & {$\mathbf{51.64}$} \\
    
    \Xhline{0.75pt}
  \end{tabular}
  \end{center}
  \caption{Tracking performance comparison on val set of WOD.
  Metrics are standard 3D MOTA and track-level recall (Recall@track) of L2 difficulty. A ground-truth object track is regarded as a track-level TP only if at least $80\%$ boxes are matched (3D IoU=$0.7$ for \textit{Vehicle}, $0.5$ for \textit{Pedestrian}) with those of a single predicted track.}
  \label{table:track-val}
\end{table}

\noindent \textbf{Cross Evaluation.}
In order to better verify the effect of our proposed principle, we reproduce the baseline tracker~\cite{ab3dmot} and motion state based object auto labeling model~\cite{3dal} and make a cross-evaluation between their modules and ours by using the same detection results (ours). In Table~\ref{table:cross-eval}, the first row can be viewed as the 3DAL approach~\cite{3dal} and achieve the lowest performance. Based on this baseline performance, using attribute-based refining modules yields $1.83$ and $1.81$ point gains for \textit{Vehicle}, $1.55$ and $1.3$ point gains for \textit{Pesdesrian}. And using offline tracking provides $1.18$ and $0.73$ point gains for \textit{Vehicle}, $0.72$ and $0.58$ point gains for \textit{Pedestrian}.
For this two groups comparison, the attribute-based refiner improves much more than offline tracker, though the object tracks are not that good, our refiner can still leverage the temporal context information. It also shows that complete object tracks are essential to affect the process of long-term sequential point clouds. The reason is revealed in Table~\ref{table:track-val}, our offline tracker yields $16.32$ point and $18.56$ point gains on Recall@track respectively, while MOTA is slightly higher. Based on the complete tracks, our attribute-based refiner could further boost the performance, as the last row of Table~\ref{table:cross-eval} and \ref{table:track-val} is shown.
These results demonstrate the strong ability of our DetZero.

\noindent \textbf{Compare with prior trackers.}
We replace our proposed offline tracker with several SOTA trackers but maintain all the other modules in our pipeline, and evaluate the trackers' performance (Recall@track) and the final performance after refining (3D APH).
The other trackers lead to degraded final APH performance because our tracker promises the completeness of tracks (Recall@track). Note that the other trackers would update the geometry size and trajectory of objects, while our offline tracker doesn't at this step.

\begin{table}[hb]
\small
\setlength{\tabcolsep}{0.11cm}
\renewcommand\arraystretch{1.1}
  \begin{center}
  \begin{tabular}{l|cc|cc}
    \Xhline{0.75pt}
      {} & \multicolumn{2}{c|}{Recall@track} & \multicolumn{2}{c}{3D APH} \\

      {} & {\textit{Vehicle}} & {\textit{Pedestrian}} & {\textit{Vehicle}} &{\textit{Pedestrian}} \\
    
    \hline
    {AB3DMOT~[\textcolor{green}{41}]} & {$23.96$} & {$26.63$} & {$78.14$} & {$78.24$} \\

    {SimpleTrack~[\textcolor{green}{23}]} & {$33.86$} & {$35.28$} & {$80.04$} & {$79.01$} \\

    {ImmortalTracker~[\textcolor{green}{39}]} & {$35.34$} & {$39.88$} & {$80.56$} & {$79.12$} \\
    
    {Ours} & 
    {$\mathbf{40.28}$} & {$\mathbf{45.19}$} & {$\mathbf{82.92}$} & {$\mathbf{81.01}$}  \\
    
    \Xhline{0.75pt}
  \end{tabular}
  \end{center}
  \caption{\footnotesize{Performance (L2) comparison on val set of WOD with different trackers.}}
\end{table}

\noindent \textbf{Generalization Ability of Refining Module.}
To better verify the generalization ability of our approach, especially the proposed attribute-based refining module, we take as input three upstream results with different qualities for inference. The \textit{low} group comes from our base detector, while the \textit{mid} group utilizes the techniques mentioned in Sec.~\ref{sec:upstream} to generate high-quality results. In \textit{high} group, we leverage the image information to further boost the upstream performance.
In Table~\ref{table:general}, our downstream refining module obtains significant improvements in all three groups. Besides, on both \textit{Vehicle} and \textit{Pedestrian}, the improvements of L2 are greater than those of L1. These results further show two strong conclusions: (1) our upstream module can recall hard samples, even if they are not over the IoU threshold of true positive, and (2) our downstream refining module takes advantage of temporal context to optimize these hard samples.

\begin{table}
\small
\setlength{\tabcolsep}{0.32cm}

\begin{center}
\renewcommand{\arraystretch}{1.1}
  \begin{tabular}{c|cc|cc}
    \Xhline{0.75pt}
     & \multicolumn{2}{c|}{\textit{Vehicle}} & \multicolumn{2}{c}{\textit{Pedestrian}} \\
    
    {} & {L1} & {L2} & {L1} & {L2} \\
    
    \hline
    {upstream (\textit{low})} & {$77.86$} & {$70.00$} & {$74.95$} & {$67.48$} \\

    \hdashline
    {+ Refine} & {$81.43$} & {$74.68$} & {$78.15$} & {$71.17$} \\

    \hdashline
    \rowcolor{blue}{\textit{improvement}} & {+$3.57$} & {+$4.68$} & {+$3.20$} & {+$3.69$} \\

    \hline
    {upstream (\textit{mid})} & {$82.57$} & {$75.24$} & {$83.07$} & {$76.34$} \\
    
    {+ Refine} & {$89.06$} & {$82.92$} & {$87.06$} & {$81.01$} \\

    \hdashline
    \rowcolor{blue}\textit{improvement} & {+$6.49$} & {+$7.68$} & {+$3.99$} & {+$4.67$} \\
    
    \hline
    {upstream (\textit{high})} & {$83.80$} & {$76.99$} & {$85.77$} & {$79.74$} \\
    
    {+ Refine} & {$89.34$} & {$83.57$} & {$88.30$} & {$82.94$} \\

    \hdashline
    \rowcolor{blue}\textit{improvement} & {+$5.54$} & {+$6.58$} & {+$2.53$} & {+$3.20$} \\
    
    \Xhline{0.75pt}
  \end{tabular}
  \end{center}
  \caption{Verifying generalization ability of our DetZero on WOD val set. Metrics are standard 3D APH of both L1 and L2 difficulties for \textit{Vehicle} and \textit{Pedestrian}.
  }
  \label{table:general}
\end{table}

\section{Conclusion}
In this work, we have proposed DetZero, a state-of-the-art offboard 3D detector using long-term sequential point clouds as input. 
The cores of our success are a multi-frame object detector and offline tracker which generates high-quality complete object tracks, and an attribute-based auto labeling model leveraging the full potential of long-term sequential point clouds.
Evaluated on WOD, our method has ranked $1$st place, showing remarkable margins over prior art 3D detectors.
Moreover, the extensive ablation studies and analysis lead to convincing evaluation and application exploration with such high-quality perception results.

\section*{Acknowledgments}
This project is funded in part by National Key R\&D Program of China Project 2022ZD0161100, by the Centre for Perceptual and Interactive Intelligence (CPII) Ltd under the Innovation and Technology Commission (ITC)’s InnoHK, by General Research Fund of Hong Kong RGC Project 14204021, and by the Science and Technology Commission of Shanghai Municipality (No. 22DZ1100102). Hongsheng Li is a PI of CPII under the InnoHK.

{\small
\bibliographystyle{ieee_fullname}
\bibliography{egbib}
}

\clearpage

\appendix
\section*{Appendix}

\section{Overview}
This document is the supplementary material of submission 4378. We provide more details of models, experiments and analysis results in this document.
Sec.~\ref{sec:app-objectdet} introduces more details about our multi-frame 3D object detectors.
Sec.~\ref{sec:app-track} describes the implementation details of our offline tracking.
In Sec.~\ref{sec:app-refiner}, the network structures and the model training details are described together with figures and tables. Sec.~\ref{sec:app-experiment} provides more experiment results and analyses. We also visualize the results after our DetZero in Sec.~\ref{sec:app-vis}.

\section{Multi-frame 3D Object Detection}
\label{sec:app-objectdet}
In this section, we provide more detailed explanations of the multi-frame 3D object detector. Please refer to Table~\ref{tab:detection} for the detailed ablation study of multi-frame detectors. Firstly, we take CenterPoint~\cite{centerpoint} as our base detector owing to producing dense detection results, which is beneficial for the downstream refine module. 

\noindent \textbf{Multi-frame Input.} We accumulate LiDAR sweeps to utilize temporal information and to densify the LiDAR point cloud. The past $4$ frames combined with the current frame serve as our input point cloud. To distinguish points from different sweeps, we also follow~\cite{ding20201st} to add a time offset as an additional attribute to the point cloud. Moreover, 3-frame input (past $2$ + current $1$) is also used to make up more detection models for boosting the performance in the subsequent model ensembling.

\noindent \textbf{Two-stage Module.} To obtain more accurate bounding boxes, we introduce the Point Density-Aware Voxel
network (PDV)~\cite{pdv} as the two-stage module to refine the coarse proposals coming from the multi-frame base detector. This model can leverage the voxel point centroid localization and account for point density variations to enhance refining features.

\noindent \textbf{Model Ensembling.}
Following~\cite{deepfusion,afdetv2}, we use different TTA settings to boost the inference performance: $\left[ 0^\circ, \pm 22.5^\circ, \pm 45^\circ, \pm 135^\circ, \pm 157.5^\circ, 180^\circ \right]$ for global rotation along z-axis, $\left[ 0.95, 1.05 \right]$ for global scaling. Besides, different grid sizes of $\left[0.075, 0.075, 0.15 \right]$m and $\left[ 0.1, 0.1, 0.15 \right]$m are used to train both $5$-frame and $3$-frame input models.
Finally, we adopt 3D version WBF~\cite{wbf} to fuse different model results combined with the above TTAs.

\noindent \textbf{Training Details.} We use Adam optimizer~\cite{dpk2015adam} with one-cycle learning rate policy, with max learning rate $3 \times 10^{-3}$, weight decay $0.01$ and momentum $0.85$ to $0.95$. We also adopt the common data augmentations including global rotation, global scaling, translation along z-axis and gt-sampling to train the base detector for $20$ epochs. The total batch size is set as $64$. The gt-sampling is removed for last $5$ epochs training~\cite{pointaugmenting}. We train another $6$ epochs for two-stage refinement without gt-sampling, while keeping the same batch size and learning rate as the first stage. Besides the general classification and regression loss functions, we also add the IoU loss function~\cite{ciassd} to better account for the center-based object detection.

\begin{figure}[ht]
    \centering
    \includegraphics[width=0.4\textwidth]{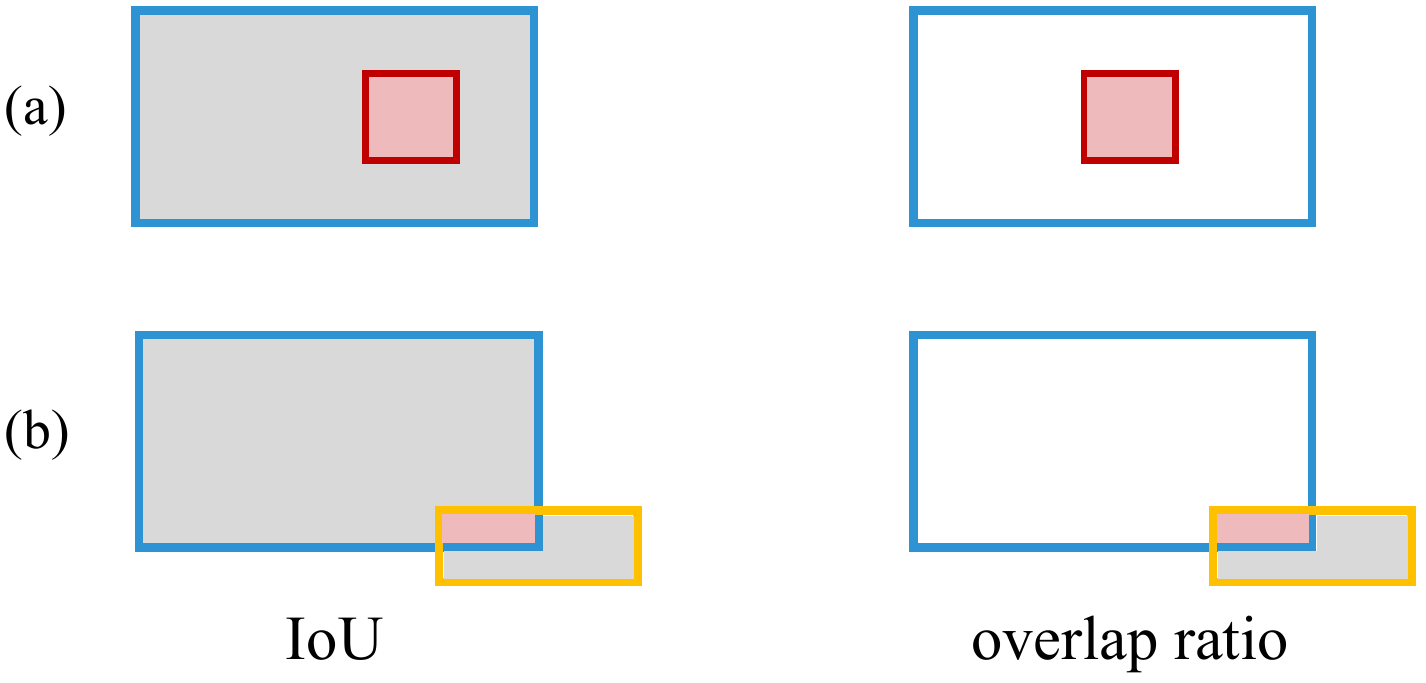}
    \caption{The comparison between IoU and overlap ratio. We show two examples here, the blue, red and orange boxes represent vehicle, pedestrian and cyclist from BEV respectively. And the gray region is the union of two boxes while the pink region is the intersection for best view. In (a), the false positive (FP) pedestrian is totally inside the vehicle, the IoU value between these two boxes is still small, while the overlap ratio equals $1$ (the denominator is the same as the numerator). In (b), the gray regions are quite different for these two methods, so the overlap ratio metric could lead to small FPs filtering.}
    \label{fig:iou}
\end{figure}

\begin{table*}
\small
\setlength{\tabcolsep}{0.38cm}
\renewcommand\arraystretch{1.1}
  \begin{center}
  \begin{tabular}{ccccc|cc}
    \Xhline{0.75pt}
    {Base detector} & {Multi-frame}&{0.075 Voxel}&{Two-stage}& {TTA} & {\textit{Vehicle} (L1 / L2)} & {\textit{Pedestrian} (L1 / L2)} \\
    \hline
    {\checkmark} & {} & &  {} & {} & {$74.51$ / $66.44$} & {$70.56$ / $63.57$}  \\

    {\checkmark} & {\checkmark} & {} &  & {} & {$78.61$ / $71.07$} & {$78.78$ / $71.46$} \\

    {\checkmark} & {\checkmark} & {\checkmark}&  & {} & {$79.57$ / $72.04$} & {$81.09$ / $73.16$}  \\

    {\checkmark} & {\checkmark} & {} &  {\checkmark} & & {$81.02$ / $73.15$} & {$80.32$ / $72.39$} \\
    
    {\checkmark} & {\checkmark} & {\checkmark} &  {\checkmark} & & {$81.17$ / $73.29$} & {$81.14$ / $74.00$} \\

    {\checkmark} & {\checkmark} & {\checkmark} & {\checkmark} & {\checkmark} & {$\mathbf{82.57}$ / $\mathbf{75.09}$} & {$\mathbf{83.23}$ / $\mathbf{76.47}$} \\
    
    \Xhline{0.75pt}
  \end{tabular}
  \end{center}
  \vspace{-2mm}
  \caption{Effect of each component in our multi-frame Detection module on WOD val set. Metrics are 3D APH of both L1 and L2 difficulties for \textit{Vehicle} and \textit{Pedestrian}. 
  }
  \label{tab:detection}
\end{table*}

\section{Implementation Details of Offline Tracking}
\label{sec:app-track}
Our multi-frame 3D detector is encouraged to generate sufficient bounding boxes. Hence, we utilize pre-processing operations to stabilize the association of our offline tracker. To be specific, we found that there are many boxes overlapped with each other. And some small boxes are even completely wrapped by other boxes, for example, the vehicle boxes contain pedestrian boxes. In this situation, the traditional IoU-based calculation will be invalid, as shown in Fig.~\ref{fig:iou}. Therefore, we adopt a new metric to determine whether a box should be kept or filtered out, which is called \textit{overlap ratio}. For each box (subject), we first calculate the pair-wise intersection area with other boxes (object), which serve as the numerator. Then, we use the original area of the object box as the denominator to get the result, and the value range is $[0, 1]$. This overlap ratio can filter out the overlapped boxes of small objects as shown in Fig.~\ref{fig:iou}. We also report the quantitative performance in Table~\ref{tab:app-tracking-recall}. In our implementation, we use BEV overlap ratio and set the thresholds as $0.3$ for \textit{Vehicle}, $0.2$ for \textit{Pedestrian} and \textit{Cyclist}.

In our two-stage data association, the high-score group contains boxes satisfying two options: (1) the confidence score is larger than $0.1$, and (2) there are more than $3$ ($3$ for \textit{Vehicle}, $1$ for \textit{Pedestrian} and \textit{Cyclist}) points inside the box. Otherwise, the boxes are assigned to the low-score group. The threshold used for association is different for the two groups. In the high-score group, the newly detected boxes are first associated with pre-existing object tracks by BEV IoU ($0.3$, $0.15$, $0.15$). The unmatched boxes are used to generate new object tracks and fed into the low-score group for next-stage association ($0.2$, $0.1$, $0.1$). After successful association, we would replace the trajectories with matched detected boxes, rather than updating them through Kalman filtering~\cite{kalman}.

In the life cycle management, the birth rate and death rate of an object track are set to $1$ and infinite. When any two object tracks overlap with each other and the ratio is larger than the threshold ($0.5$, $0.4$, $0.4$), we will merge them together by keeping the earlier birth object ID. Afterward, any redundant boxes that have not been updated are removed.

\section{Implentation Details of Attribute-based Refining}
\label{sec:app-refiner}
In this section, we provide the details of the network structure, training strategies and loss functions of each refining model.

\begin{figure}[ht]
    \centering
    \includegraphics[width=0.42\textwidth]{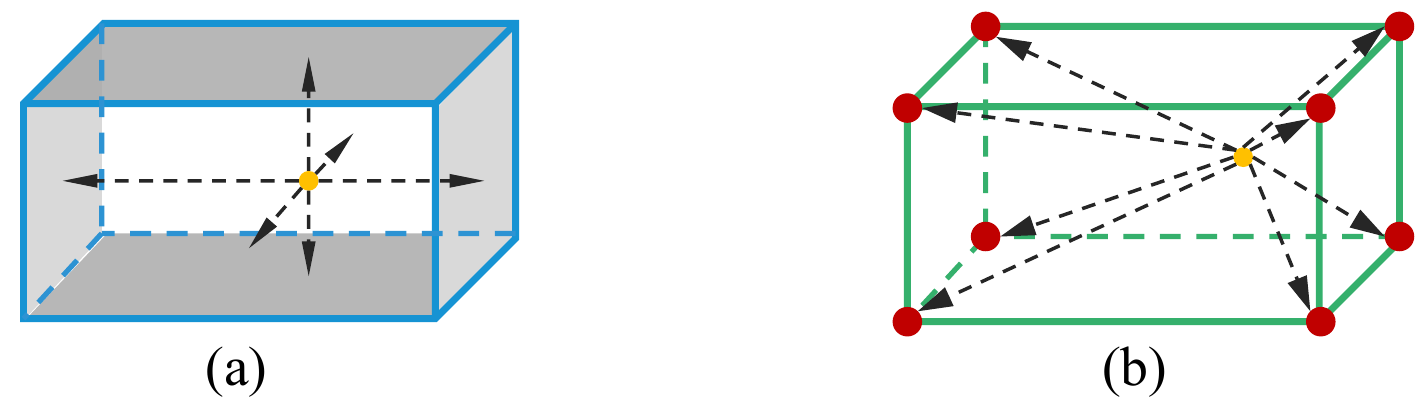}
    \caption{The point-to-surface encoding in GRM (a) and point-to-corner encoding in PRM (b). All the distances are three dimensional ($x, y, z$). Note that there are a few points outside the corresponding proposal box.}
    \label{fig:point-to-proposal}
\end{figure}

\subsection{Geometry Refining Model}
\noindent \textbf{Encoder Network Structures.}
In our GRM, the query encoder and value encoder are both PointNet~\cite{pointnet} structured. Each layer is built as a multi-layer perceptron (MLP) followed by batch normalization and ReLU activation layer. The query encoder ENC$_1$ takes as input $t$ randomly selected samples to generate corresponding geometry queries $\mathbf{Q}^\text{geo} \in \mathbb{R}^{t \times D}$. Meanwhile, the selected $n$ geometry-aware points (after proposal-to-surface encoding shown in Fig.~\ref{fig:point-to-proposal}) are fed into the value encoder ENC$_2$ to generate the global point feature, serving as $\mathbf{V}^\text{geo} \in \mathbb{R}^{n \times D}$.
The details of point cloud processing are shown in Table~\ref{tab:geo-query-enc} and Table~\ref{tab:geo-value-enc} respectively.

\begin{table}[hb]
\setlength{\tabcolsep}{6pt}
\begin{center}
\begin{tabular}{|c|c|c|c|}
\hline
{Index} &{Input} &{Operation} & {Output Shape}\\ \hline
(1) & - & geometry points $f^\text{geo}$ & $t \times 256 \times 11 $ \\
(2) & (1) & Linear$(11 \rightarrow 128 )$ & $t \times 256 \times 128$ \\
(3) & (2) & ReLU, BN & $t \times 256 \times 128$ \\
(4) & (3) & Linear$(128 \rightarrow 128)$ & $t \times 256 \times 128$ \\
(5) & (4) & ReLU, BN & $t \times 256 \times 128$ \\
(6) & (5) & Linear$(128 \rightarrow 256)$ & $t \times 256 \times 256$ \\
(7) & (5) & Max pooling & $t \times 256$ \\
(8) & (7) & Linear$(256 \rightarrow 256)$ & $t \times 256$ \\
(9) & (8) & ReLU, BN & $t \times 256$
\\
\hline
\end{tabular}
\vspace{0.5mm}
\caption{The architecture of query encoder in GRM. $t$ is the number of randomly selected proposals for each object track. For each proposal, we randomly sample $256$ points.}
\label{tab:geo-query-enc}
\end{center}
\end{table}

\begin{table}[hb]
\setlength{\tabcolsep}{6pt}
\begin{center}
\begin{tabular}{|c|c|c|c|}
\hline
{Index} &{Input} &{Operation} & {Output Shape}\\ \hline
(1) & - & geometry points $f^\text{geo}$ & $n \times 10 $ \\
(2) & (1) & Linear$(10 \rightarrow 128 )$ & $n \times 128$ \\
(3) & (2) & ReLU, BN & $ n \times 128 $ \\
(4) & (3) & Linear$(128 \rightarrow 128)$ & $n \times 128$ \\
(5) & (4) & ReLU, BN & $n \times 128$ \\
(6) & (5) & Linear$(128 \rightarrow 512)$ & $n \times 512$ \\
(7) & (5) & Max pooling & $128$ \\
(8) & (7) & Repeat & $n \times 128$ \\
(9) & (5)(8) & Concatenate & $n \times 640$
\\
(10) & (9) & Linear$(640 \rightarrow 256)$ & $n \times 256$ \\
(11) & (10) & ReLU, BN & $n \times 256$
\\
\hline
\end{tabular}
\vspace{1mm}
\caption{The architecture of value encoder in our GRM.}
\label{tab:geo-value-enc}
\end{center}
\end{table}

\noindent \textbf{Attention-based Decoder.}
Our decoder layer follows the classical design, which consists of a multi-head self-attention layer, a multi-head cross-attention layer and an FFN with residual structure. We adopt 1-layer structure in our implementation.

For the multi-head self-attention layer ($\mathbf{SA}$), we enrich contextual relationships and feature differences among selected samples. Specifically, we map the object queries $\mathbf{Q}^\text{geo}$ by linear projections $\mathbf{W}_1, \mathbf{W}_2, \mathbf{W}_3$ to form the so-called query, key, and value. For simplicity, we omit the superscript ``geo''. Then, the output after $\mathbf{SA}$ is given by
\begin{equation}
\label{eq:self-attn}
\footnotesize
    \mathbf{SA}\left( \mathbf{Q}^\text{geo} \right) = \left[ \sum_{m=0}^{t} { \frac{\exp{\left(\mathbf{W}_1 \mathbf{q}_i \left(\mathbf{W}_2 \mathbf{q}_m \right)^T \right)}}{\sum_{j=0}^{t} \exp{\left(\mathbf{W}_1 \mathbf{q}_j \left(\mathbf{W}_2 \mathbf{q}_j \right)^T \right)}} \mathbf{W}_3 \mathbf{q}_m} \right] 
\end{equation}
where $[\cdot]$ is a concatenation operation and $\mathbf{Q}^\text{geo}$ can be divided into $\left[ \mathbf{q}_1, \cdots, \mathbf{q}_i, \cdots, \mathbf{q}_t \right], i = 1, ..., t$.

For the multi-head cross-attention layer ($\mathbf{CA}$), the refined object queries can aggregate relevant context from global point features for compensating supplementary views. And the calculation is expressed by
\begin{center}
\tiny
    \begin{equation}
    \label{eq:cross-attn}
    \mathbf{CA}\left( \mathbf{Q}^\text{geo}, \mathbf{K}^\text{geo}, \mathbf{V}^\text{geo} \right) = \left[ \sum_{m=0}^{t} { \frac{\exp{\left(\mathbf{W}_4 \mathbf{q}_i \left(\mathbf{W}_5 \mathbf{k}_m \right)^T \right)}}{\sum_{j=0}^{t} \exp{\left(\mathbf{W}_4 \mathbf{q}_j \left(\mathbf{W}_5 \mathbf{k}_j \right)^T \right)}} \mathbf{W}_6 \mathbf{v}_m} \right] 
    \end{equation}
\end{center}
where $\mathbf{W}_4, \mathbf{W}_5, \mathbf{W}_6$ are linear projections, $\mathbf{K}^\text{geo}$ can be divided into $\left[ \mathbf{k}_1, \cdots, \mathbf{k}_t \right]$, and $\mathbf{V}^\text{geo}$ can be divided into $\left[ \mathbf{v}_1, \cdots, \mathbf{v}_t \right]$.

\noindent \textbf{Training Details.}
During training, we randomly selected $t=3$ object proposals as geometry queries. While in inference, the 3 samples are selected with the highest scores. For each query, we predict its size classes (among pre-defined template size classes) and residual sizes for each size class. 
The size classes are supervised with a cross-entropy loss $L_\text{cls}$ while the residual sizes are supervised with a L1 loss $L_\text{reg}$. The total geometry refining loss is $L^\text{geo} = 0.1 L_\text{cls} + 2 L_\text{reg} $.
The final geometry size is the average of these 3 predictions, which is then assigned to all the frames of the object track.

The randomly selected $n=4096$ geometry-aware points are augmented through randomly flipping along X, Y axes with 50\% chance, and randomly rotating around the Z-axis by Uniform$\left[ -\frac{\pi}{2}, \frac{\pi}{2} \right]$ degrees, and randomly scaling by Uniform$[0.9, 1.1]$. During inference, we also adopt TTA settings, in which the scaling operation can boost the performance at most while the flipping and rotation along z-axis operations lead to slight improvements.

We use Adam optimizer with a one-cycle decay policy to separately train the model for each class. The initial learning rate is $0.001$ and the batch size is set to $128$. The total epochs are $30$ for \textit{Vehicle}, $100$ for \textit{Pedestrian} and $500$ for \textit{Cyclist}. In total, we have extracted around $44$K vehicle tracks, $18$K pedestrian tracks and $0.5$K cyclist tracks for training. Ground-truth boxes are assigned to every frame of the object track (frames with no matched ground-truth are skipped, such as the non-point objects).

\subsection{Position Refining Model}

\noindent \textbf{Encoder Network Structures.}
The encoders in PRM are similar to those in GRM. Each object track is padded with zeros to the length of the whole sequence, such as $200$ for WOD. The full processing procedures are shown in Tabel~\ref{tab:prm-query-enc}.

\begin{table}[t]
\begin{center}
\begin{tabular}{|c|c|c|c|}
\hline
{Index} &{Input} &{Operation} & {Output Shape}\\ \hline
(1) & - & position points $f^\text{geo}$ & $200 \times 256 \times 32 $ \\
(2) & (1) & Linear$(32 \rightarrow 128 )$ & $200 \times 256 \times 128$ \\
(3) & (2) & ReLU, BN & $200 \times 256 \times 128$ \\
(4) & (3) & Linear$(128 \rightarrow 128)$ & $200 \times 256 \times 128$ \\
(5) & (4) & ReLU, BN & $200 \times 256 \times 128$ \\
(6) & (5) & Linear$(128 \rightarrow 256)$ & $200 \times 256 \times 256$ \\
(7) & (6) & Max pooling & $200 \times 256$ \\
(8) & (7) & Linear$(256 \rightarrow 256)$ & $200 \times 256$ \\
(9) & (8) & ReLU, BN & $200 \times 256$
\\
\hline
\end{tabular}
\vspace{0.5mm}
\caption{The architecture of query encoder in PRM. The object track is padded to the length of $200$. For each proposal of the object track, we randomly sample $256$ points.}
\label{tab:prm-query-enc}
\end{center}
\end{table}

\noindent \textbf{Attention-based Decoder.}
The full attention-based process is designed to model the local-to-global position contextual relations, which is similar as mentioned in Eq.~\ref{eq:self-attn} and Eq.~\ref{eq:cross-attn}.

\noindent \textbf{Training Details.}
The object tracks whose length are short than $7$ are deprecated during training, and we also adopt a random frame deprecation as the additional data augmentation. Random flipping operation could boost the performance at most by stabilizing the trajectories during inference.
The residual distances between each tracked box to the randomly-selected proposal's center are supervised with an L1 loss $L_\text{reg}^\text{ce}$. For heading prediction, we also utilize a bin-based classification and residual degrees. We use $12$ heading anchors, each bin accounts for $30$ degrees from $0$ to $360$ degrees. The total loss is $L^\text{pos} = L_\text{reg}^\text{ce} + 0.1 L_\text{cls}^\text{yaw} + 2 L_\text{reg}^\text{yaw}$.
We train total $50$ epochs for \textit{Vehicle} with a batch size of $96$, $100$ epochs for \textit{Pedestrian} with a batch size of $128$, and $200$ epochs for \textit{Cyclist} with a batch size of $64$. The optimizer setting is the same as our GRM.

\subsection{Confidence Refining Model}
For the first classification branch of our CRM, we use different IoU thresholds to determine the positive and negative samples. Specifically, $\tau_h$ is set to $0.7$ for \textit{Vehicle}, $0.5$ for \textit{Pedestrian} and \textit{Cyclist}, $\tau_l$ is set to $0.35$, $0.25$ and $0.25$ respectively. We use binary labels $0$ and $1$ for supervision. For both two branches, we use BCE loss to supervise the predictions. And the total loss $L^\text{conf} = L_\text{cls} + L_\text{iou}$. We train total $30$ epochs for \textit{Vehicle} with a batch size of $256$, $50$ epochs for \textit{Pedestrian} with a batch size of $256$, and $100$ epochs for \textit{Cyclist} with a batch size of $64$. The optimizer setting is the same as our GRM.

\section{Details of the Human Label Study}
We keep the same setting as 3DAL~\cite{3dal}, and the randomly selected 5 sequences from the WOD val set are listed in Table~\ref{tab:human-seq}. We directly utilize their human labeling results rather than repeat the whole labeling task. In summary, there are 12 experienced labels to annotate the 15 labeling tasks (3 sets of re-labels for each sequence) and obtain 2.3k labels. Then, the human APs are computed by comparing them with the WOD’s released ground-truth labels and using the number of points in boxes as human label scores.

\begin{center}
\begin{table}[hb]
\setlength{\tabcolsep}{6pt}
\begin{tabular}{|c|}
\hline
{Sequence} \\ \hline
segment-17703234244970638241\_220\_000\_240\_000 \\
segment-15611747084548773814\_3740\_000\_3760\_000 \\
segment-11660186733224028707\_420\_000\_440\_000 \\
segment-1024360143612057520\_3580\_000\_3600\_000 \\
segment-6491418762940479413\_6520\_000\_6540\_000 \\
\hline
\end{tabular}
\vspace{1mm}
\caption{The list of selected sequences from WOD val set for human label study.}
\label{tab:human-seq}
\end{table}
\end{center}

Besides, we also report the statistical results of auto labels (on 90\% train set) in Table~\ref{tab:human-statistic} to better show that the auto labels contain fewer boxes than ground-truth, especially for the hard cases (object points are smaller than 5). Therefore, the student model trained with auto labels would generate fewer false positives than with GT labels. Besides, when we remove the boxes by cutting different scores, the student model trained with auto labels can preserve more true positive boxes, which proves that the model is more confident in the easy samples. We infer that the model can focus more on the easy samples with better convergence.

\begin{table}[hb]
\begin{center}
\footnotesize
\setlength{\tabcolsep}{0.08cm}
\renewcommand{\arraystretch}{1.1}
  \begin{tabular}{c|cc|cc}
    \Xhline{0.75pt}
     & \multicolumn{2}{c|}{\textit{Vehicle}} & \multicolumn{2}{c}{\textit{Pedestrian}} \\
    
    {} & {$\ge$ 5 pts} & {$<$ 5 pts} & {$\ge$ 5 pts} & {$<$ 5 pts} \\

    \hline
    {Ground-truth} & {$3.44$} & {$0.50$} & {$1.54$} & {$0.30$} \\

    {Auto labels} & {$2.87$} & {$0.17$} & {$1.29$} & {$0.16$} \\
    
    \Xhline{0.75pt}
  \end{tabular}
  \end{center}
  \vspace{-2mm}
  \caption{The comparison between ground-truth and auto labels. Boxes in auto labels have an IoU larger than thresholds ($0.7$ for \textit{Vehicle} and $0.5$ for \textit{Pedestrian}) would be kept for statistics in millions.}
  \label{tab:human-statistic}
\end{table}

\section{More Experiments}
\label{sec:app-experiment}

\noindent \textbf{Comparison on different distances.}
To better evaluate the effect of our DetZero, we report the performance on different distances. As shown in Tabel~\ref{tab:app-distance}, for both \textit{Vehicle} and \textit{Pedestrian}, the improvements are increasing while the distances are from near to far. It proves that the current performance bottleneck of object detection exists at the farther range. And our DetZero could utilize the long-term temporal context to optimize these boxes located at the beginning and the end of an object track. In addition, the improvements of objects with L2 difficulty are larger than those of L1 difficulty, which draws the same conclusion as Table~\ref{table:general}.

\begin{table}[t]
\begin{center}
\footnotesize
\setlength{\tabcolsep}{0.08cm}
\renewcommand{\arraystretch}{1.1}
  \begin{tabular}{c|cc|cc|cc|cc}
    \Xhline{0.75pt}
     & \multicolumn{2}{c|}{\textit{Total}} & \multicolumn{2}{c|}{\textit{0-30m}} & \multicolumn{2}{c|}{\textit{30-50m}} & \multicolumn{2}{c}{\textit{50+m}} \\
    
    {} & {L1} & {L2} & {L1} & {L2} &  {L1} & {L2} &  {L1} & {L2} \\
    
    \hline
    {Upstream} & {$82.57$} & {$75.24$} & {$94.25$} & {$93.35$} & {$81.54$} & {$75.25$} & {$63.28$} & {$51.32$} \\

    {Full} & {$89.06$} & {$82.92$} & {$96.27$} & {$95.52$} & {$88.41$} & {$83.97$} & {$77.80$} & {$65.70$} \\

    \rowcolor{blue}{\textit{improve}} & {+$6.49$} & {+$7.68$} & {+$2.02$} & {+$2.17$} & {+$6.87$} & {+$8.72$} & {+$\mathbf{14.52}$} & {+$\mathbf{14.38}$} \\

    \hline
    {Upstream} & {$83.07$} & {$76.34$} & {$86.09$} & {$82.26$} & {$82.09$} & {$75.39$} & {$77.35$} & {$65.22$} \\

    {Full} & {$87.06$} & {$81.01$} & {$89.25$} & {$85.71$} & {$86.14$} & {$80.84$} & {$83.08$} & {$71.94$} \\

    \rowcolor{blue}{\textit{improve}} & {+$3.99$} & {+$4.67$} & {+$3.16$} & {+$3.45$} & {+$4.05$} & {+$5.47$} & {+$\mathbf{5.73}$} & {+$\mathbf{6.72}$} \\
    
    \Xhline{0.75pt}
  \end{tabular}
  \end{center}
  \vspace{-2mm}
  \caption{Performance evaluation of different distances on WOD val set. Metrics are standard 3D APH of both L1 and L2 difficulties for \textit{Vehicle} (first group) and \textit{Pedestrian} (second group).
  }
  \label{tab:app-distance}
\end{table}

\noindent \textbf{Offline tracking generates complete tracks.}
We list the top SOTA tracking methods on Waymo 3D tracking leaderboard~\footnote{We report the performance of 3D detection and tracking till 2023-03-08 23:59 GMT.} in Table~\ref{tab:tracking-leaderboard}. Our DetZero ranks 1st place by outperforming previous SOTA performance with $9.97$-point MOTA (L2) for all classes. Compared to our own upstream results, we still keep a huge performance improvement with $5.84$-point MOTA (L2) for all classes.

We also show the effect of generating sufficient complete object tracks in Table~\ref{tab:app-tracking-recall}. The first row shows that the detection results contain huge false-positive boxes, resulting in very low precision performance. Traditional IoU-based filtering operations will loose the effect when facing overlapped boxes. As a comparison, our overlap ratio based filtering would further remove these boxes, especially under a loose threshold. Finally, the whole offline tracking procedure would further remove FPs while keeping a slightly-low Recalls.

\begin{table}
\begin{center}
\footnotesize
\setlength{\tabcolsep}{0.08cm}
\renewcommand{\arraystretch}{1.1}
  \begin{tabular}{c|cc|cc}
    \Xhline{0.75pt}
     & \multicolumn{2}{c|}{\textit{Vehicle} (0.7 / 0.5)} & \multicolumn{2}{c}{\textit{Pedestrian} (0.5 / 0.3)} \\
    
    {} & {Recall} & {Precision} & {Recall} & {Precision} \\

    \hline
    {Detection} & {$\mathbf{83.6}$ / $\mathbf{95.6}$} & {$13.4$ / $15.3$} & {$\mathbf{88.9}$ / $\mathbf{97.1}$} & {$6.7$ / $7.3$} \\

    {IoU filer} & {$75.3$ / $92.7$} & {$52.6$ / $65.1$} & {$83.8$ / $95.0$} & {$17.8$ / $20.2$} \\

    {OR filer} & {$75.2$ / $92.4$} & {$55.8$ / $69.0$} & {$82.7$ / $93.5$} & {$20.7$ / $23.4$} \\

    {Offline Trk.} & {$75.4$ / $91.8$} & {$\mathbf{66.2}$ / $\mathbf{81.9}$} & {$81.2$ / $91.3$} & {$\mathbf{35.7}$ / $\mathbf{40.3}$} \\
    
    \Xhline{0.75pt}
  \end{tabular}
  \end{center}
  \vspace{-2mm}
  \caption{Performance comparison of our offline tracking. Metrics are 3D Recall and Precision under different IoU thresholds for \textit{Vehicle} (0.7 / 0.5) and \textit{Pedestrian} (0.5 / 0.3). OR filter is the filtering operation based on the overlap ratio.
  }
  \label{tab:app-tracking-recall}
\end{table}

\noindent \textbf{Effect of point cloud information encoding.}
We show the ablation of point cloud information encoding methods used in GRM and PRM. For every experiment, we randomly selected 20\% sequences (160) of the original train set for training, and evaluate the performance on the whole val set (202 sequences). We also report the Accuracy performance with the object's motion state, which is calculated by its ground-truth trajectory. In Table~\ref{tab:grm-abl}, our point-to-surface (\textit{p2s}) encoding method yields the largest gains. In Table~\ref{tab:prm-abl}, the point-to-corner (\textit{p2co}) encoding method yields the largest gains compared to point-to-center (\textit{p2ce}) encoding. Because the tracked boxes have already provided efficient geometry information, which could be efficiently utilized by our encoding method. We also find that the improvements after position refining are much higher than those after geometry refining, which further demonstrates the effect of our PRM on removing jitters and smoothing trajectories through attending global motion information.

\begin{table}[hb]
\footnotesize
\setlength{\tabcolsep}{0.15cm}
\renewcommand\arraystretch{1.1}
  \begin{center}
  \begin{tabular}{ccc|cc|c|c}
    \Xhline{0.75pt}
      \multirow{2}{*}{\textit{xyzi}} & \multirow{2}{*}{\textit{p2s}} & \multirow{2}{*}{\textit{score}} & \multicolumn{2}{c|}{ALL} & {Static} & {Dynamic} \\
    
     &  &  & {box} & {track} & {box} & {box} \\
    
    \hline

    {\checkmark} & {} & {} & 
    {$78.08$} & {$66.87$} & 
    {$76.18$} & {$84.12$} \\

    {\checkmark} & {\checkmark} & {} & 
    {$78.50$} & {$67.36$} & 
    {$76.60$} & {$84.43$} \\
    
    {\checkmark} & {\checkmark} & {\checkmark} & 
    {$\mathbf{78.56}$} & {$\mathbf{67.42}$} & 
    {$\mathbf{76.66}$} & {$\mathbf{84.51}$} \\
    
    \Xhline{0.75pt}
  \end{tabular}
  \end{center}
  \vspace{-2mm}
  \caption{Effect of the different point encoding method for GRM. Metrics are Accuracy under standard IoU ($0.7$ for \textit{Vehicle}) for both box-level and track-level statistics. We split the objects based on its ground-truth motion state for better comparison.}
  \label{tab:grm-abl}
\end{table}

\begin{table}[hb]
\footnotesize
\setlength{\tabcolsep}{0.15cm}
\renewcommand\arraystretch{1.1}
  \begin{center}
  \begin{tabular}{cccc|cc|c|c}
    \Xhline{0.75pt}
      \multirow{2}{*}{\textit{xyzi}} & \multirow{2}{*}{\textit{p2ce}} & \multirow{2}{*}{\textit{p2co}} & \multirow{2}{*}{\textit{score}} & \multicolumn{2}{c|}{ALL} & {Static} & {Dynamic} \\
    
     &  &  & & {box} & {track} & {box} & {box} \\
    
    \hline

    {\checkmark} & {} & {} & {} & 
    {$78.95$} & {$68.78$} & 
    {$78.36$} & {$80.81$} \\

    {\checkmark} & {\checkmark} & {} & {\checkmark} & 
    {$80.98$} & {$71.95$} & 
    {$80.71$} & {$81.84$} \\

    {\checkmark} & {\checkmark} & {\checkmark} & {} & 
    {$81.84$} & {$72.83$} & 
    {$81.30$} & {$83.55$} \\
    
    {\checkmark} & {\checkmark} & {\checkmark} & {\checkmark} & 
    {$\mathbf{81.99}$} & {$\mathbf{73.22}$} & 
    {$\mathbf{81.47}$} & {$\mathbf{83.60}$} \\
    
    \Xhline{0.75pt}
  \end{tabular}
  \end{center}
  \vspace{-2mm}
  \caption{Effect of the different point encoding method for PRM. Metrics are Accuracy under standard IoU ($0.7$ for \textit{Vehicle}) for both box-level and track-level statistics. We split the objects based on its ground-truth motion state for better comparison.}
  \label{tab:prm-abl}
\end{table}

\noindent \textbf{Effect of the number of geometry queries.}
We show the empirical performance by selecting different object samples as geometry queries. As shown in Table~\ref{tab:grm-query}, the performance increases while the number of queries increases, which could be viewed as another data augmentation method. Note that the performance gaps among them may not be very stable and we finally select $3$ queries in our whole processing.

\begin{table}[hb]
\footnotesize
\setlength{\tabcolsep}{0.15cm}
\renewcommand\arraystretch{1.1}
  \begin{center}
  \begin{tabular}{c|cc|c|c}
    \Xhline{0.75pt}
      \multirow{2}{*}{query number} & \multicolumn{2}{c|}{ALL} & {Static} & {Dynamic} \\
    
     & {box} & {track} & {box} & {box} \\
    
    \hline

    {1} & 
    {$78.29$} & {$66.81$} & 
    {$76.38$} & {$84.27$} \\

    {2} & 
    {$78.48$} & {$67.23$} & 
    {$76.46$} & {$84.34$} \\
    
    {3} & 
    {$78.56$} & {$\mathbf{67.42}$} & 
    {$\mathbf{76.66}$} & {$\mathbf{84.51}$} \\

    {5} & 
    {$\mathbf{78.57}$} & {$67.32$} & 
    {$\mathbf{76.66}$} & {$84.50$} \\
    
    \Xhline{0.75pt}
  \end{tabular}
  \end{center}
  \vspace{-2mm}
  \caption{Effect of the different number of geometry queries used in GRM. Metrics are Accuracy under standard IoU ($0.7$ for \textit{Vehicle}) for both box-level and track-level statistics. We split the objects based on its ground-truth motion state for better comparison.}
  \label{tab:grm-query}
\end{table}

\section{Qualitative Results}
\label{sec:app-vis}
\begin{figure*}
\centering
\includegraphics[width=0.95\textwidth]{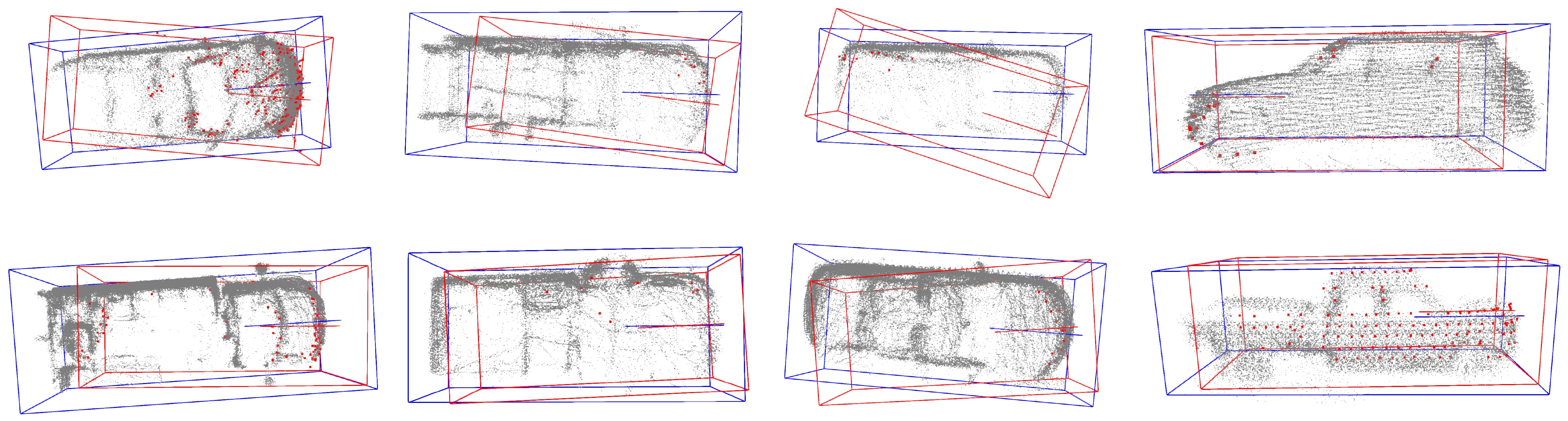}
\caption{The visualization of GRM results on WOD val set. The red boxes are selected from one frame of the object track, and corresponding points are also colored with red. The refining boxes with precise sizes are colored with blue.}
\label{fig:geo-vis-res}
\end{figure*}

\begin{figure*}
\centering
\includegraphics[width=1.\textwidth]{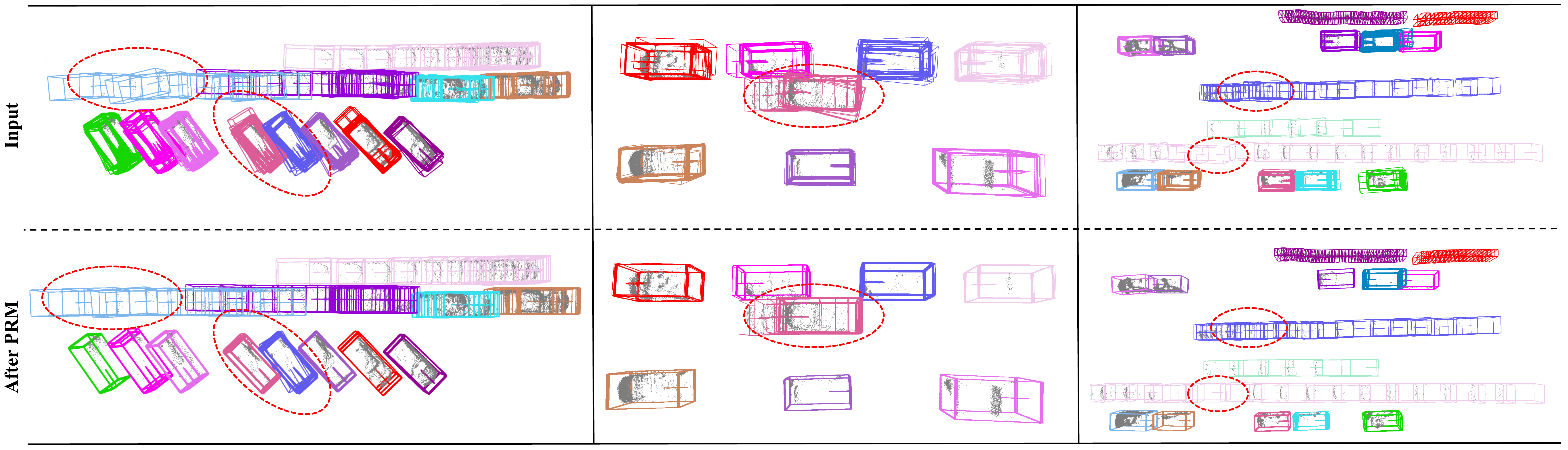}
\caption{The visualization of PRM results on WOD val set. The first row is the input object tracks, and the second row is the corresponding results after PRM. We use red dotted circles to mark the important cases.}
\label{fig:pos-vis-res}
\end{figure*}

In this section, we show the qualitative comparisons after our attribute-refining module in Fig.~\ref{fig:geo-vis-res} and Fig.~\ref{fig:pos-vis-res}. Please refer to our website for more video visualizations.

\clearpage

\begin{table*}[]
\footnotesize
\setlength{\tabcolsep}{0.09cm}
\renewcommand\arraystretch{1.1}

  \begin{center}
  \begin{tabular}{l|c|c|c|cc|cc|cc}
    \Xhline{0.75pt}
    \multirow{2}{*}{ Method } & \multirow{2}{*}{Rank} & \multirow{2}{*}{Frames} & MOTA & \multicolumn{2}{c|}{\textit{Vehicle} (MOTA $\uparrow$ /MOTP$\downarrow$)} & \multicolumn{2}{c|}{\textit{Pedestrian} (MOTA $\uparrow$ /MOTP$\downarrow$)} & \multicolumn{2}{c}{\textit{Cyclist} (MOTA $\uparrow$ /MOTP$\downarrow$)} \\

    {} & {} & {} & L2 & L1 & L2 & L1 & L2 & L1 & L2 \\

    \hline
    $\text{DetZero}$ (Full) & $1$ & $200$ & $\mathbf{75.05}$ & $\mathbf{79.04}$ / $\mathbf{14.09}$ & $\mathbf{75.97}$ / $\mathbf{14.18}$ & $\mathbf{77.60}$ / $\mathbf{28.76}$ & $\mathbf{76.03}$ / $\mathbf{28.76}$ & $\mathbf{73.24}$ / $\mathbf{23.77}$ & $\mathbf{73.16}$ / $\mathbf{23.77}$ \\

    $\text{DetZero}$ (Upstream) & $-$ & $200$ & $69.21$ & $71.02$ / $15.47$ & $67.96$ / $15.47$ & $71.56$ / $29.90$ & $70.00$ / $29.90$  & $69.75$ / $24.24$  & $69.67$ / $24.24$  \\
    
    \hline
    $\text{InceptioLidar}^*$ & $2$ & $10$ & $65.08$ & $68.78$ / $15.68$ & $65.58$ / $15.70$ & $66.38$ / $29.54$ & $64.52$ / $29.54$ & $65.19$ / $25.42$ & $65.12$ / $25.42$ \\
    $\text{HorizonMOT3D}$~\cite{HorizonMOT3D} & $3$ & $5$ & $63.45$ & $67.30$ / $15.75$ & $64.07$ / $15.77$ & $65.88$ / $30.67$ & $64.15$ / $30.67$ & $62.20$ / $25.45$ & $62.13$ / $25.45$ \\

    $\text{MFMS\_Track}^*$ & $4$ & $4$ & $63.27$ & $66.45$ / $15.65$ & $63.14$ / $15.65$ & $65.47$ / $30.19$ & $63.85$ / $30.19$ & $62.90$ / $25.44$ & $62.83$ / $25.44$ \\

    $\text{CasTrack}^*$ & $5$ & $5$ & $62.60$ & $66.95$ / $15.79$ & $63.66$ / $15.79$ & $66.39$ / $30.22$ & $64.79$ / $30.24$ & $59.41$ / $25.30$ & $59.34$ / $25.30$ \\

    $\text{ImmortalTracker}$~\cite{immortal_tracker} & $6$ & $2$ & $60.92$ & $63.77$ / $16.22$ & $60.55$ / $16.22$ & $62.20$ / $31.17$ & $60.60$ / $31.20$ & $61.68$ / $27.41$ & $61.61$ / $27.41$ \\

    $\text{OptMOT}^*$ & $7$ & $2$ & $60.85$ & $65.47$ / $16.16$ & $62.18$ / $16.16$ & $60.02$ / $30.58$ & $58.31$ / $30.58$ & $62.14$ / $26.97$ & $62.06$ / $26.97$ \\

    $\text{SimpleTrack}$~\cite{simpletrack} & $8$ & $2$ & $60.18$ & $63.53$ / $16.19$ & $60.30$ / $16.23$ & $61.75$ / $31.09$ & $60.13$ / $31.14$ & $60.18$ / $27.35$ & $60.12$ / $27.35$ \\

    $\text{CenterPoint}$~\cite{centerpoint} & $11$ & $2$ & $58.67$ & $62.58$ / $16.30$ & $59.38$ / $16.37$ & $58.28$ / $31.13$ & $56.64$ / $31.16$ & $60.06$ / $27.62$ & $60.00$ / $27.62$ \\

    \Xhline{0.75pt}
  \end{tabular}
  \end{center}
  \vspace{-2mm}
  \caption{Performance comparison on the Waymo 3D tracking leaderboard. Metrics are standard 3D MOTA and MOTP by both L1 and L2 difficulties.
  Anonymous submissions are marked with $*$.
  }
  \label{tab:tracking-leaderboard}
\end{table*}

\end{document}